\documentclass[review,3p]{elsarticle}

\usepackage{hyperref}

\journal{Journal of \LaTeX class}

\usepackage{amsmath,amssymb,amsfonts}
\usepackage{graphicx}
\usepackage{textcomp}
\def\BibTeX{{\rm B\kern-.05em{\sc i\kern-.025em b}\kern-.08em
    T\kern-.1667em\lower.7ex\hbox{E}\kern-.125emX}}

\usepackage{booktabs}
\usepackage{enumitem}
\usepackage{color}
\usepackage{graphics}
\usepackage{multirow} 
\usepackage{pifont}
\usepackage{float}
\usepackage{bbm} 
\usepackage{mathtools} 
\usepackage{url}
\usepackage{bm}

\usepackage{subfigure}
\usepackage{caption,setspace}

\usepackage{xspace}
\newcommand*{\eg}{e.g.\@\xspace}
\newcommand*{\ie}{i.e.\@\xspace}
\newcommand*{\etal}{\emph{et al.}\@\xspace}

\begin{document}

\begin{frontmatter}

\title{Accuracy vs. Complexity: A Trade-off in Visual Question Answering Models}

\author[add1,add2]{Moshiur R. Farazi\corref{mycorrespondingauthor}}
\cortext[mycorrespondingauthor]{Corresponding author}
\ead{moshiur.farazi@anu.edu.au}
\author[add3,add1]{Salman H. Khan}
\author[add1]{Nick Barnes}

\address[add1]{College of Engineering and Computer Science, Australian National University (ANU), Canberra ACT 0200 Australia}
\address[add2]{Data61, Commonwealth Scientific and Industrial Research Organisation (CSIRO), Canberra ACT 2601 Australia}
\address[add3]{Inception Institute of Artificial Intelligence (IIAI), Abu Dhabi 0000 UAE}

\begin{abstract}
Visual Question Answering (VQA) has emerged as a Visual Turing Test to validate the reasoning ability of AI agents. 
The pivot to existing VQA models is the joint embedding that is learned by combining the visual features from an image and the semantic features from a given question. 
Consequently, a large body of literature has focused on developing complex joint embedding strategies coupled with visual attention mechanisms to effectively capture the interplay between these two modalities.
However, modelling the visual and semantic features in a high dimensional (joint embedding) space is computationally expensive, and more complex models often result in trivial improvements in the VQA accuracy.
In this work, we systematically study the trade-off between the model complexity and the performance on the VQA task.
VQA models have a diverse architecture comprising of pre-processing, feature extraction, multimodal fusion, attention and final classification stages. 
We specifically focus on the effect of ``multi-modal fusion" in VQA models that is typically the most expensive step in a VQA pipeline.
Our thorough experimental evaluation leads us to two proposals, one optimized for minimal complexity and the other one optimized for state-of-the-art VQA performance. 
\end{abstract}

\begin{keyword}
Visual question answering, Convolutional neural networks, Recurrent neural networks, Multi-modal fusion, Speed-accuracy trade-off
\end{keyword}

\end{frontmatter}


\section{Introduction}
\label{sec:intro}
The Visual Question Answering (VQA) problem aims to a develop a deep understanding of both vision and language, and the complex interplay between the two, such that a machine is able to answer intelligent questions about a visual scene. The VQA task is inspired by the astounding ability of humans to perceive and process information from multiple modalities and draw connections between them. An AI agent equipped with VQA ability has wide applications in service robots, personal digital assistants, aids for visually impaired and interactive educational tools, to name a few \cite{antol2015vqa, gu2018recent}. 

Given the success of deep learning, one common approach to address the VQA problem is by extracting visual features from an input image or a video using pretrained Convolutional Neural Networks (CNNs) \eg, VGGNet \cite{simonyan2014very}, ResNet \cite{he2016deep}, ResNeXt \cite{xie2017aggregated}; and representing language features from the input questions using Recurrent Neural Networks (RNN) \eg, \cite{antol2015vqa, fukui2016multimodal, kiros2015skip}. The automatic and generalized feature learning capability of deep neural networks has paved the way towards joint processing of multiple modalities in a single framework, leading to dramatic improvements on the challenging VQA task \cite{antol2015vqa,krishna2016visual,zhu2016visual7w}. 

To effectively capture the interaction between visual and semantic domains, one must learn a joint representation common between the two domains. 
Capturing the multimodal interaction between these two modalities is computationally expensive (both in terms of compute and memory footprint), especially when the interactions are learned on high-dimensional visual and language features extracted using deep neural networks. Different multimodal operations ranging from simple linear summation and concatenation to complex bilinear pooling and tensor decomposition have been proposed to effectively model this bi-modal interaction and achieve state-of-the-art VQA accuracy \cite{fukui2016multimodal,benyounescadene2017mutan, ben2019block}. 

In this paper, we specifically focus on studying the trade-off between the complexity and performance offered by different multi-modal fusion mechanisms in VQA models. The multi-modal fusion component is often the most computationally expensive part in a VQA pipeline. It is therefore of interest to analyze its impact on the final performance. Notably, VQA pipelines are often coupled with multi-level, multi-directional attention mechanisms \cite{lu2016hierarchical,yang2016stacked,jabri2016revisiting,xu2015show, kim2018bilinear} that allow the VQA model to identify most salient regions/phrases in the given image/question required to predict a correct answer. Here, we do not analyse different attention mechanisms since they are model-specific and therefore less generalizable across models and different tasks requiring multi-modal integration. However, using a simple attention approach, we demonstrate that attention is helpful in VQA settings across different fusion strategies.



The main contribution of our paper are as follows:
\begin{itemize}
    \item We provide a succinct survey of the state-of-the-art VQA models employing multimodal fusion to learn a joint embedding, and describe how most of the leading models leverage a similar high-level architecture.
    \item We establish a VQA baseline that supports the three most popular meta-architectures (visual features extractor, bilinear fusion and co-attention) and a unified evaluation protocol by varying these meta-architectures. 
    \item We perform an extensive evaluation on three challenging VQA datasets (\ie, VQAv2, VQA-CPv2 and TDIUC) for different combinations of feature extractor, bilinear fusion model and attention mechanism to generate an accuracy vs. complexity trade-off curves. 
    \item Our finding suggests VQA models using visual features obtained by Squeeze-and-excitation Network (SeNet \cite{hu2018squeeze}) mostly outperform models using widely adopted ResNet \cite{he2016deep} features, irrespective of  attention and fusion mechanism. Further, 
    we report that employing MFH fusion facilitates achieving a superior performance over its counterparts.
    \item We propose a combination of feature extractor and meta-architecture that achieves state-of-the-art performance on three most challenging VQA datasets.
\end{itemize}



\begin{figure}[t]
\centering
  \includegraphics[width=0.9\linewidth]{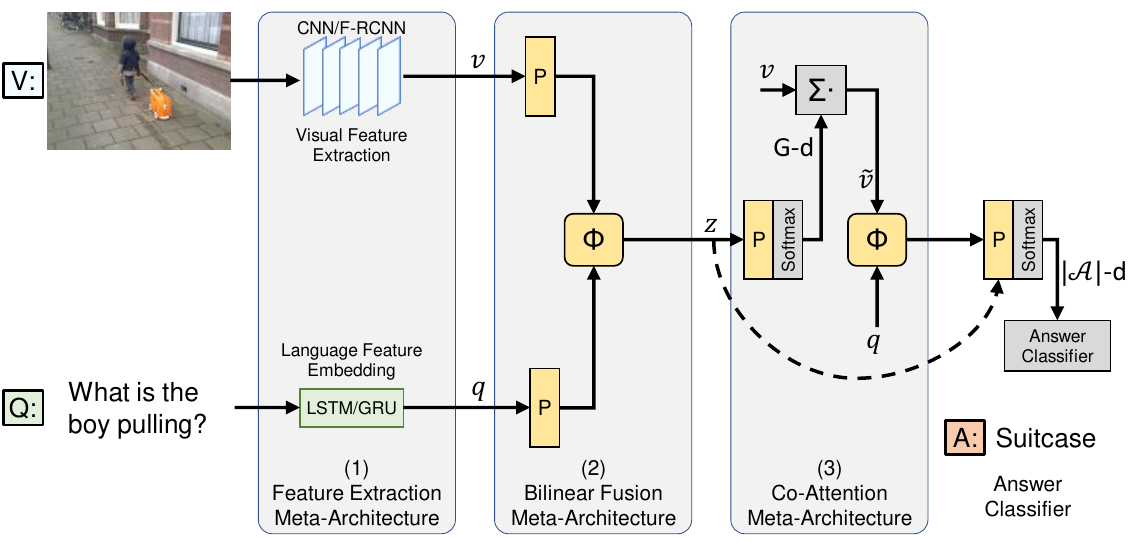}
  \caption{An unified VQA model with three components that co-occur in existing models (we term them \emph{meta-architectures}). (1) The feature extraction meta-architecture generates visual feature $\bf{v}$ and semantic feature $\bf{q}$ from the input image and question respectively. (2) The extracted features are projected to a common-space through $P$ and jointly embedded into $z$ with a  bilinear fusion model $\Phi$. (3) The attention meta-architecture takes the joint embedding feature $z$ and learn a spatial attention distribution to generate an attended visual feature representation $\tilde{v}$. The question embedding $q$ and $\tilde{v}$ are again jointly embedded and passed to the answer classifier. The joint embedding feature $z$ can be directly passed to the answer classifier to predict the answer $a^*$ skipping the co-attention meta-architecture (denoted by the dashed line). The trainable blocks are color coded yellow.  }
\label{fig:all_meta_arch}
\end{figure}

\section{VQA Model Architecture}
\label{sec:meta_architecture}
The VQA task is modeled as a classification task.  Since there exists a long tail distribution of answers in the large-scale VQA datasets,  the most frequent answers are placed in a candidate answer set $\mathcal{A}$. The goal is to predict the best possible answer $a^*$ for a natural language question $ \bf{Q} $ about an image $\bf{I}$. This can be formulated as:
\begin{equation}
\label{eq:vqa_formulation}
    a^* = \operatorname*{arg\,max}_{a \in \mathcal{A}} p (a|\mathbf{I},\mathbf{Q}; \bm{\theta})
\end{equation}
where $\bm{\theta}$ denotes the model parameters.

\begin{figure}[t]
\centering
  \includegraphics[width=0.9\linewidth]{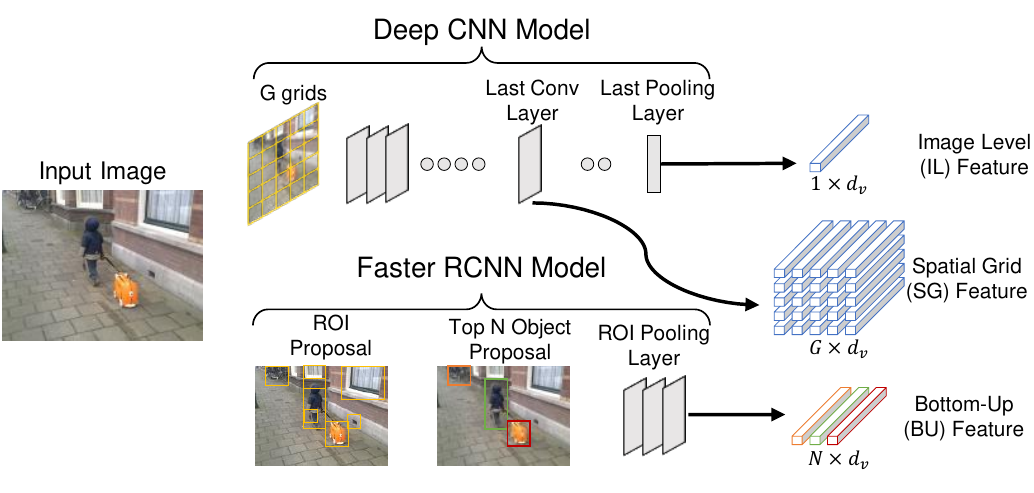}
  \caption{Visual feature extraction meta-architecture illustrating the pipeline for generating Image Level(IL), Spatial Grid (SG) and Bottom-Up(BU) from the input image.}
\label{fig:vis_feat_meta}
\end{figure}

\subsection{Feature Extraction Meta-Architecture}
\label{sec:feat_extract}
The feature extraction component consists of two parts. \emph{First,} a visual feature extraction block takes the input image and extracts visual features. \emph{Second,} a language embedding block generates a semantic embedding from the input features. As these two parts require a trained image and language model on large-scale datasets, these blocks are part of the data pre-processing done before training the VQA model itself.

\textbf{Visual Feature.} To generate discriminative features from images, similar to other high level visual reasoning tasks (\eg, image captioning, visual dialog and relationship prediction), VQA models employ deep neural networks pretrained for object recognition and detection. These deep CNN models generate a feature representation of the image $\mathbf{I}$, denoted as $\bm{v}$. It can be formulated as:
\begin{equation}
\bm{v} = \textnormal{CNN}(\mathbf{I}) = \{\bm{v}_i, \; s.t., \, i \in [1,G]\},
\end{equation}
where $\bm{v}_i \in \mathbb{R}^{d_v}$ is the feature vector of $\mathnormal{\mathnormal{i}^{th}}$ image location and $\mathnormal{G}$ is the total number of image locations in a grid. The dimension of $d_v$ and $G$ depend on how the features are extracted using a particular CNN model. The extracted visual features can be categorized into three main types (see Fig.~\ref{fig:vis_feat_meta}):
\begin{enumerate}[label=\roman*)]
    \item \textbf{Image Level (IL):} These features are extracted from the last pooling layers before the classification layer (\eg, `pool5' layer of ResNet\cite{he2016deep}). IL features are $1 \times \mathnormal{d_v}$ dimensional as they represent the features of the whole image (\ie, $\mathnormal{G}=1$). When only these features are available, the additional visual attention (discussed in Sec.~\ref{sec:attention}) is not used as these features have no spatial information.
    \item \textbf{Spatial Grid (SG):} The Spatial Grid (SG) features are extracted from the last convolutional layer (\eg, `res5c' if using ResNet-152). The spatial grid feature is $\mathnormal{G \times d_v}$ dimensional where each feature map corresponds to a uniform grid location on the input image. While using SG features, an additional attention mechanism is often used to generate a more refined visual representation based on the input question (Sec.~\ref{sec:attention}). 
    \item \textbf{Bottom-Up (BU):} Anderson \etal \cite{Anderson2017up-down} proposed to use features maps of different object proposals instead of IL or SG features. The object proposals are obtained by passing each input image through an object detector (\eg, Faster--RCNN \cite{ren2015faster}) that is pretrained 
    on large-scale object detection datasets. The extracted features are $G \times d_v$ dimensional, where $G=N$ is the number of top object proposals in an image.
\end{enumerate}

\textbf{Language Feature.} The question is first tokenized into words and encoded in to word embeddings using a pretrained sentence encoder (\eg, GLoVe \cite{pennington2014glove}, Skip-thought \cite{kiros2015skip}). The length of the word embedding is set to $l$, determined from the question length distribution in the dataset, where unusually longer question are clipped and short question are zero padded to get a fixed-sized word embedding $\bm{w}_l$. The word embeddings are passed through LSTMs \cite{hochreiter1997long} (or its variants) to obtain the semantic features $\bm{q}$ from the input question:
\begin{equation}
    \bm{q} = \textnormal{LSTM} (\bm{w}_l),
\end{equation}
where $\bm{q}$ is the output feature of the last word from the LSTM network and is of $d_q$ dimension. The dimension of the semantic feature embedding is determined by size of the hidden state of the LSTM unit. In the scope of this paper, we use a fixed language feature extraction meta-architecture for all our experiments since our goal is to study the trade-off provided by multi-modal fusion strategies. However, a more advanced word embedding, such as Bidirectional Encoder Representations from Transformers (BERT) \cite{devlin2018bert} can provide additional performance gains.

\subsection{Fusion Model Meta-Architecture}
\label{sec:fusion}
The second meta-architecture (common to VQA models) jointly embeds the extracted visual and semantic features into a common space. To this end, a multimodal embedding function $\Phi$ is learned:
\begin{equation}
    \bm{z} = \Phi(\bm{q},\bm{v})
\end{equation}
where $\bm{z}$ is the learned joint embedding from the input question and image. The simplest way to project $\bm{q}$ and $\bm{v}$ into the same space is by taking Hadamard product of the inputs, $\bm{z} = \bm{v} \odot \bm{q}$. However, this operation requires the inputs to be of equal dimension and is limited to a linear model. 


To fuse visual and semantic features of equal or different dimension and capture the complex interaction between these two modalities, one can adopt bilinear fusion models and take the outer product of the two input feature vectors:
\begin{equation}
\label{eq:bilinear_model}
    \bm{z} = \mathcal{W} [\bm{v} \otimes \bm{q}]
\end{equation}
where $\mathcal{W} \in \mathbb{R}^{d_v \times d_q \times \mathcal{A}_d}$ is the learned fusion model, $\mathcal{A}_d$ is the number of entries in the candidate answers set, while $\otimes$  and $\mathnormal{[~]}$ denote outer product and vectorization operations, respectively. This operation allows the model $\mathcal{W}$ to learn the interactions between the inputs in a multiplicative manner. One major limitation of this approach is $\mathcal{W}$ is very high dimensional. For example, if one is using SG features with a ResNet-152 backbone, LSTM with $2048$ hidden dimensions and an answer classifier with $3000$ candidate answers, the leaned model $\mathcal{W}_i$ for $i^{\textnormal{th}}$ image grid location will be $\mathbb{R}^{2048 \times 2048 \times 3000}$. As a result, even with a simplistic design, the VQA model will have over $12$ billion learnable parameters which is expensive both computationally and memory-wise. Several models have been proposed to tackle this problem and in this paper we aim to investigate the trade-off between complexity and accuracy of VQA models by using a variety of bilinear fusion methodologies. 

We first establish two simple baseline multi-modal models and then formulate different bilinear models proposed in the literature in our experimental setting. The baseline models we experiment with in the scope of this work are as follows. 

\textbf{Linear:} Linear Summation is the simplest multi-modal fusion model that we experiment with. It first transforms the input feature vector into an intermediate space through fully connected layers. The intermediate features are then added together and projected back to the answer prediction space through another fully connected layer. This operation only uses a linear operation (\ie, summation) in the intermediate space to capture the interaction between the visual and language features, thus is dubbed Linear.

\textbf{C-MLP:} The second baseline fusion model is called Concatenation-MLP (C-MLP). We first concatenate the input features in their native space and pass the resulting visual-semantic features through a 3 layer Multi-layer Perceptron (MLP). The MLP model learns to non-linearly encode the concatenated features and produces a joint-embedding feature in the answer prediction space. 

\textbf{MCB:} Multi-modal Compact Bilinear (MCB) pooling \cite{fukui2016multimodal} introduced the use of bilinear models to perform fusion between visual and semantic feature vectors in a VQA setting. First, the input feature vectors are approximated as $\bm{q}'$ and $\bm{v}'$ by using count-sketch projection \cite{charikar2002finding} and then their element wise product is taken in the spectral domain. The spectral domain transformation is achieved via a Fast Fourier Transform (FFT).
\begin{equation}
\label{eq:mcb}
    \bm{z} = \textnormal{FFT}^{-1} (\textnormal{FFT}(\bm{v}') \odot \textnormal{FFT}(\bm{q}')).
\end{equation}
This operation leverages the property that convolution in the time domain is equivalent to element-wise product in the frequency domain; and the frequency domain product is converted back to the original domain by an inverse Fourier transformation. However, the model is still quite expensive to train as it requires the resultant joint embedding vector $z$ to be high-dimensional (precisely $16,000d$) to have a superior VQA accuracy.

\textbf{MLB:} To reduce the dimensions of the output feature vector, Kim \etal \cite{kim2016hadamard} proposed Multimodal Low-rank Bilinear Pooling (MLB). MLB uses a low-rank factorization of input features vectors during bilinear operation. The input feature vectors, $\bm{v}$ and $\bm{q}$ are projected to a joint embedding space $z \in \mathbb{R}^{d_z}$ and their Hadamard product is taken as follows:
\begin{equation}
\label{eq:mlb}
    \bm{z} = (P_v^T \bm{v}) \otimes (P_q^T \bm{q}) 
\end{equation}
where $P_v$ and $P_q$ are projection matrices of dimension $\mathbb{R}^{d_v \times d_z}$ and $\mathbb{R}^{d_q \times d_z}$ respectively. Here, the output joint embedding size $d_z$ is set to $~1,000$. Generally, the VQA model thus developed achieves better accuracy than the former MCB approach.

\textbf{MFB:} Even though MLB achieves comparable performance with MCB, it takes longer to converge. Multi-modal Factorized Bilinear Pooling (MFB) \cite{yu2018beyond} proposed to add a pooling operation on the jointly embedded feature vector. This process is divided in two stages. First, during the \emph{Expand} stage, the projection dimension is expanded by a factor $k$ and the input visual feature vectors are projected onto $k \times d_z$ dimension. Second, during the \emph{Squeeze} stage, a sum-pooling operation is performed with size $k$ of non-overlapping windows, which squeezes the joint feature embedding by the same factor $k$.
\begin{equation}
\label{eq:mfb}
    \bm{z} = \textnormal{Sum-Pool} \big( (\tilde{P}_v^T \bm{v}) \otimes (\tilde{P}_q^T \bm{q}) , ~ k \big)
\end{equation}
where the new projection matrices are denoted by $\tilde{P}_v \in \mathbb{R}^{d_v \times d_z \times k}$ and $\tilde{P}_q \in \mathbb{R}^{d_q \times d_z \times k}$. After sum-pooling over $k$ windows, the joint embedding feature vectors again become $d_z$ dimensional. It can be see that setting $k=1$, MLB can be considered as a spacial case of MFB. The inclusion of the sum-pooling operation with a factor $k$ improves the convergence of the VQA model and provides boost in VQA accuracy compared to MLB.

\textbf{MFH:} To model a more complex interactions, Multi-modal Factorized High-order pooling (MFH) \cite{yu2018beyond} uses a series of cascading MFB blocks. Each MFB block takes the input feature vectors and internal feature of the previous MFB block. The internal feature of $i^{th}$ MFB block among a total of $m$ cascaded MFB blocks can be formulated as:
\begin{equation}
\label{eq:mfh1}
    \bm{z}_{\textnormal{int}}^{i} = 
    \begin{cases}
    \mathbbm{1} \otimes \big((\tilde{P}_v^T \bm{v}) \otimes (\tilde{P}_q^T \bm{q})\big),& \textnormal{when } i = 1\\
    z_{\textnormal{int}}^{i-1} \otimes \big((\tilde{P}_v^T \bm{v}) \otimes (\tilde{P}_q^T \bm{q})\big),& \textnormal{when } i > 1
    \end{cases},
\end{equation}
where $i \in [1,m]$ and $\mathbbm{1}$ is a $d_z \times k$ dimensional matrix of all ones. $\bm{z}_{\textnormal{int}}^i \in \mathbb{R}^{d_z  k}$ is similar to the output of the \emph{Expand} stage of MFB except for the additional input from the previous MFB block. The output joint embedding of the $i^{th}$ MFB block is formulated as:
\begin{equation}
\label{eq:mfh2}
    \bm{z}^i = \textnormal{Sum-Pool} (\bm{z}_{\textnormal{int}}^{i}) 
\end{equation}
Finally, the final output of a MFH operation with $m$ MFBs block is obtained by concatenating the output of each MFB block:
\begin{equation}
\label{eq:mfh3}
    \bm{z} = [\bm{z}^1,\bm{z}^2, ... \bm{z}^m].
\end{equation}
Here, the output joint embedding vector $\bm{z} \in \mathbb{R}^{d_z m}$. When $m=1$, then MFB can be considered as a spacial case of MFH.

\textbf{Mutan:} Multimodal Tucker Fusion (Mutan) \cite{benyounescadene2017mutan} first proposed tensor decomposition techniques to reduce the dimensionality of input visual and semantic feature vectors, and the output joint feature embedding in a VQA model. We can re-write Eq.~\ref{eq:bilinear_model} to obtain joint embedding vector $\bm{z}$ with tensor notation as:
\begin{equation}
\label{eq:bilinear_model_tensor}
    \bm{z} = (\mathcal{W} \times_1 \bm{v}) \times_2 \bm{q},
\end{equation}
where the operator $\times_i$ defines $i^{\mathnormal{th}}$ mode product between the learned  tensor $\mathcal{W}$ and input feature vectors. Following Tucker decomposition \cite{tucker1966some}, the 3-way learned model tensor $\mathcal{W}$ can be decomposed into a core tensor and three factor matrices:
\begin{equation}
\label{eq:mutan1}
    \mathcal{W} \coloneqq \mathcal{T}_c \times_1 F_v \times_2 F_q \times_3 F_z
\end{equation}
with the core tensor $\mathcal{T}_c \in \mathbb{R}^{d_{pv} \times d_{pq} \times d_z}$, and visual, question and joint embedding factor matrices are respectively $F_v \in \mathbb{R}^{d_v \times d_{pv}}$, $F_q \in \mathbb{R}^{d_q \times d_{pq}}$ and $F_z \in \mathbb{R}^{|\mathcal{A}| \times d_{z}}$. The factor matrices $F_v$, $F_q$ project the input feature vectors to $d_{pv}$ and $d_{pq}$ dimensional space, respectively, and the core tensor $\mathcal{T}$ models the interaction between the projected feature vectors and the output joint embedding. Now, to encode the fully bilinear interaction  in the joint embedding space $z$, we can formulate Eq.~\ref{eq:bilinear_model_tensor} as:
\begin{equation}
\label{eq:mutan2}
    z = (\mathcal{T}_c \times_1 F_v^T \bm{v}) \times_2 F_q^T \bm{q}. 
\end{equation}
Here, the dimensions $d_{pv}$ and $d_{pq}$ directly contribute to the model complexity and are usually set to $\sim 300$ with $d_z$ set to $\sim 500$. Comparing MLB (Eq.~\ref{eq:mlb}) and Mutan (Eq.~\ref{eq:mutan2}), MLB can be considered as a spacial case of Mutan if $d_{pv} = d_{pq} = d_z$ and the core tensor $\mathcal{T}_c$ is set to identity. This approach is more efficient compared to MLB as the rank of the core tensor is constrained which balances the interaction between the input feature vectors to achieve a higher accuracy.

\textbf{Block:} In Mutan, the multimodal interaction is solely modelled by the core tensor $\mathcal{T}_c$ which captures the rich interaction between the input features but is limited by the dimensions of the output joint embedding space. This causes the VQA accuracy to saturate for a given setting of intermediate projection dimension. To overcome this bottleneck, a block-superdiagonal tensor based decomposition (Block) technique for VQA was proposed by \cite{ben2019block}. The 3-way learned model tensor $\mathcal{W}$ is decomposed in $n$ blocks/chunks as follows:
\begin{equation}
\label{eq:block1}
    \mathcal{W} = \mathcal{T}_B \times_1 F_v \times_2 F_q \times_3 F_z,
\end{equation}
where $\mathcal{T}_B \in \mathbb{R}^{(d_{pv} n) \times (d_{pq} n) \times (d_z n)}$ and $F_v = [F_v^1, F_v^2, ... , F_v^n]$ (with a similar formulation for $F_q, F_z$). Each of the $n$ core tensor blocks represents bilinear interaction between chunks of input features. Dividing the core tensor and its factor matrices into blocks allows the model to capture the interaction between several chunks of input feature vectors that get mapped into the joint embedding space. The joint embedding feature output of $i^{\textnormal{th}}$ block is:
\begin{equation}
\label{eq:block2}
    z^i =  (\mathcal{T}_b^i \times_1 {(F_v^i)}^T \bm{v}^i) \times_2 {(F_q^i)}^T \bm{q}^i ,
\end{equation}
where $i \in [1,n]$ and the dimensions of $i^{\textnormal{th}}$ core tensor and other factor matrices are reduced by a factor of $n$ compared to the same variables in Eq.~\ref{eq:block1}. The final output joint embedding feature vector $\bm{z}$ is computed as the concatenation of $n$ block term joint embedding features as:
\begin{equation}
    \label{eq:block3}
    \bm{z} = [\bm{z}^1, \bm{z}^2, ... , \bm{z}^n]
\end{equation}
where $\bm{z} \in \mathbb{R}^{d_z}$. If we set, $n=1$ in Eq.~\ref{eq:block3}, meaning only one core tensor is used to model the interaction between the input features, block-superdiagonal tensor based decomposition becomes the Tucker decomposition as in Eq.\ref{eq:mutan1}.



\subsection{Attention-based Meta-Architecture}
\label{sec:attention}
Different questions about the same image would require a VQA model to attend to different spatial regions within an image. An additional attention mechanism allows the VQA models to identify relevant image regions for answering the question by learning an attention distribution. As mentioned in Sec.~\ref{sec:feat_extract}, each location of the SG and BU features represent a spatial grid location or an object proposal, respectively. This visual attention can be applied to a Spatial Grid (SG) and/or Bottom-Up (BU) image features where $G > 1$, where an attention mechanism learns to identify which grid locations or object proposals are most relevant in answering the given question. This question specific attention generally allows the model to achieve superior performance.

In the attention meta-architecture, we experiment with \emph{co-attention}. The co-attention process consists of two steps each of which requires the model to learn a joint-embedding feature vector from visual and semantic features (see component (3) in Fig.~\ref{sec:meta_architecture}).  In the \textbf{first} step, the model learns to generate an attention distribution vector 
using the input visual and language features. Irrespective of the bilinear embedding module used, the model learns an attention probability distribution $\alpha \in \mathbb{R}^G$ for input visual features with $G > 1$ spatial/object locations:
\begin{equation}
\label{eq:att1}
    \alpha = \textnormal{Softmax} \big( P_{\alpha} \sigma (\mathcal{W}[\bm{v} \otimes \bm{q} \cdot \mathbbm{1} ]) \big),
\end{equation}
where $\mathbbm{1}$ denotes the repeat (tile) operation to make the question feature $d_q \times G$ dimensional, $P_{\alpha} \in \mathbb{R}^{d_z \times G}$ projects the joint embedding features to $G$ dimensions and $\sigma$ is a non-linear activation function (usually $tanh$ or sigmoid). It has been found (\cite{fukui2016multimodal, yu2017multi}) that learning multiple attention distributions, commonly termed as glimpse, increases the VQA accuracy. At each glimpse $t$, the models learns an attention distribution $\alpha^t$ that results in a better  probability distribution.

In the \textbf{second} stage, the attention distribution ${\alpha}^t$ is used to take a weighted sum of the input visual features in $G$ spatial locations. The attended visual feature for glimpse $t$ can be formulated as:
\begin{equation}
\label{eq:att2}
    \tilde{v}^t = \sum_{g=1}^G \alpha^t \bm{v},
\end{equation}
where $\tilde{\bm{v}}^t \in \mathbb{R}^{d_v}$. If $t>1$, attended visual features over multiple glimpses are concatenated as $\tilde{\bm{v}} = [\bm{v}^1, \bm{v}^2, ... \bm{v}^{t_{f}}]$, where $t_f$ is the last glimpse. The final attended visual feature representation undergo a second bilinear embedding with the question feature:
\begin{equation}
\label{eq:att3}
    p(a|\tilde{\bm{v}},\bm{q};\bm{\theta}) = \textnormal{Softmax} \big( P_{|\mathcal{A}|} \sigma (\mathcal{W}[\bm{v} \otimes \bm{q}]) \big),
\end{equation}
where $P_{|\mathcal{A}|} \in \mathbb{R}^{d_z \times d_{|\mathcal{A}|}}$ is a projection matrix to the candidate answer space, $p$ is the posterior probability distribution in that space and $\bm{\theta}$ denotes the same parameter set as described in Eq.~\ref{eq:vqa_formulation}.

\section{An Unified VQA Model}
As discussed in the previous section, the VQA model is made of three main components. Different state-of-the-art models use different combinations of these meta-architectures to achieve superior performance. To experiment with different extracted features and bilinear models, we first establish a modular Unified VQA (U-VQA) architecture that supports different variations of the meta-architectures.

\textbf{Visual Feature Extractor:} We extract IG and SG visual features using the following pre-trained deep CNN models\footnote{https://github.com/Cadene/pretrained-models.pytorch} for object detection:
\begin{itemize}
    \item Inception Net \cite{szegedy2015going}: Several versions of Inception net have been proposed over the years. In our experiments, we used the InceptionNet-V4 with $d_v=1536$ and $G=12\times12$. This means the IG features are $1536$ dimensional and SG features are $1536 \times 12 \times 12$. Compared to other visual features extracted with pretrained object detectors, Inception features are the lowest dimensional visual features that we experiment with.
    \item ResNet \cite{he2016deep}: Visual features extracted by ResNet are widely used in a VQA setting. In our experiment, we use the Facebook implemented version of ResNet-152\footnote{https://github.com/facebookarchive/fb.resnet.torch} model, which has a slightly better performance compared to the original ResNet implementation. IG and SG features extracted with ResNet152 are $2048$ and $2048 \times 14 \times 14$ dimensional, respectively.
    \item ResNext \cite{xie2017aggregated}: ResNext reported better performances than the  counterpart ResNet architecture on the ImageNet and COCO detection datasets. We used ResNext101 in our experiments  and the extracted IG and SG features have same dimension as ResNet features.
    \item SeNet \cite{hu2018squeeze}: In complement to spatial features, SeNet adaptively re-calibrates the channel-wise features to achieve a higher accuracy om ImageNet dataset. We use the SeNet154 model in our experiments which also has the same IG and SG feature dimensions as ResNet152 and ResNext101.
    \item PolyNet \cite{zhang2017polynet}: We use PolyNet to extract IG and SG features which are $2048$ and $2048 \times 12 \times 12$ dimensional, respectively. The $d_v$ dimension of PolyNet features are equal to ResNet, ResNext and SeNet, but it has fewer spatial grid locations compared to the former models. 
\end{itemize}

We use the BU features\cite{Anderson2017up-down} made available by their official online repository\footnote{https://github.com/peteanderson80/bottom-up-attention}. The BU features are extracted by using a Faster-RCNN model with a  ResNet-101 backbone for the top $N$ object proposals for each image. $N$ can be adaptive (top 10 to 100 proposals) or fixed (top 36 proposals). For our experiments we use the BU features with $N=36$.

\textbf{Language Feature Embedding:} 
Similar to \cite{ben2019block}, we use a pretrained Skip-thought\cite{kiros2015skip} vectors and GRUs to encode the language features. 
The language feature embedding is set to $d_q = 2400$ for all our experiments.

\textbf{Multi-modal Fusion Models:} To embed the multi-modal features into a joint embedding space, we experiment with fusion models discussed in Sec.~\ref{sec:fusion} and two additional baseline fusion models, namely `Linear' and `C-MLP'. Following are the hyper parameters settings that we use for experimenting with these fusion models:
\begin{itemize}
    \item Linear: The intermediate dimension where the visual and semantic features are projected is set to $1000$. The $1000$ dimensional features are summed and projected to the candidate answer space.  
    \item C-MLP: The visual and question features are concatenated and passed through a MLP layer with $1600$ hidden dimensions. The output dimension of the MLP is set to the dimension of the candidate answer space.
    \item MCB: We set the joint embedding size to $16,000$ following the original implementation details reported in \cite{fukui2016multimodal}.
    \item MLB: The joint embedding size $d_z$ is set to $1200$ following the original implementation details reported in \cite{kim2016hadamard}.
    \item MFB: Following the notation in Sec.~\ref{sec:fusion}, $k$ is set to 5 and $d_z$ is set to $1000$ following the original implementation.
    \item MFH: For MFH, we keep the values of $k$ and $d_z$ same as the values used in the  MFB implementation, and cascade size is set $m=2$ MFB blocks.
    \item Mutan: Following the notation described in Sec.~\ref{sec:fusion}, we restrict the rank $\mathcal{T}_c$ to $10$ and $d_z$ is set to $700$.
    \item Block: The rank of block core tensor $\mathcal{T}_B$ is set to 15, $d_z$ is set to $1600$ and the number of blocks/chunks $n$ is set to $18$ following the original implementation. 
\end{itemize}
In our experiments, we use the official implementation of Mutan and Block, and PyTorch implementation of MCB from \url{https://github.com/Cadene/block.bootstrap.pytorch}. We re-implement MLB, MFB and MFH bilinear models in our unified VQA architecture in PyTorch\cite{NEURIPS2019_9015}.

\textbf{Co-Attention Mechanism:} We learn an attention distribution map on the SG or BU features by using a co-attention mechanism. The learned attention probability distribution $\alpha$ indicates which spatial grid locations (for SG features) or object proposals (for BU features) are more important for answering the input question. For all our experiments, we use two glimpse, which means for a given image-question pair, two different $\alpha$ are generated. These attention distribution maps are applied on the input visual features separately and the resulting visual representation is concatenated to a vector of size $2 \times d_v$.

\section{Datasets}
We perform extensive evaluation on three VQA benchmark datasets, namely VQAv2 \cite{Goyal_2017_CVPR}, VQA-CPv2 \cite{Agrawal_2018_CVPR} and TDIUC \cite{kafle2017analysis}. The first dataset we experiment on is \textbf{VQAv2}\cite{Goyal_2017_CVPR}. This dataset is a refined version of the VQAv1 \cite{antol2015vqa} dataset as it introduces complementary image-question pairs to mitigate the language bias present in the original dataset. The VQAv2 dataset contains over $204K$ images from the MSCOCO dataset \cite{lin2014microsoft} and $1.1M$ open-ended questions paired with these images (an average of 6 questions per image). Each question has $10$ ground-truth answers sourced from crowd-workers for the open-ended questions. The evaluation on the VQAv2 test-set can only be done by submitting the evaluation file in their online evaluation server, and offline evaluation can be done by training the model on train split and evaluating the models performance on the validation split. For this reason, we report validation scores on Tab.~\ref{tab:VQAv2_valset} and the best test-standard scores in Tab.~\ref{tab:VQAv2_SOTA}. 

Further, we experiment on the Visual Question Answering under Changing Priors (VQA-CP) dataset. The \textbf{VQA-CPv2} dataset is re-purposed from the training and validation sets of the VQAv2 dataset. Similar Image-Question-Answer(IQA) triplets of the training and validation splits of the VQAv2 dataset are grouped together and then re-distributed into train and test sets in a way that questions within the same question type (\eg, `what color', `how many' etc.) and similar ground-truth answers are not repeated in test and train splits. This makes it harder for any VQA model to leverage the language bias to artificially achieve a higher accuracy. 

Finally, we perform experiments on the Task Directed Image Understanding Challenge (TDIUC) dataset. The \textbf{TDIUC} dataset divides the VQA paradigm into 12 different task directed question types. These include questions that require a simpler task (e.g., object presence, color attribute) and more complex tasks (e.g., counting, positional reasoning). The IQA triplets are sourced from train and validation set of VQA dataset and Visual Genome \cite{krishna2016visual} dataset, but undergo some automatic and manual annotations to generate the ground-truth.

\textbf{Evaluation Metric:}
While experimenting on the VQA and VQA-CP dataset, we report VQA accuracy following the standard protocol \cite{antol2015vqa,Goyal_2017_CVPR,krishna2016visual}. The accuracy of a predicted answer $a^*$ is:
\begin{equation}
\label{eq:vqa_acc}
\textnormal{VQA Accuracy} = \min \bigg(\frac{\# \; \text{of humans answered } a^*}{3}, 1\bigg)
\end{equation}
which means if the predicted answer $a^*$ is given by at least 3 human annotators out of 10, then it will be considered correct. We report overall accuracy on the dataset for all question types along with `Yes/No', `Number' and `Other' question types.

When evaluating on the TDIUC dataset, we report accuracy for each of the 12 question types defined in the dataset. It allows us to further evaluate the capacity of a fusion model to answer diverse types of questions that require different reasoning capabilities. As the VQA datasets are crowd-sourced, they have an inherent bias due to a skewed question distribution. Along with the individual accuracy, we also report arithmetic and harmonic means across all per-question-type accuracy, dubbed arithmetic mean-per-type (Arithmetic MPT) and harmonic mean-per-type accuracy (Harmonic MPT). Arithmetic MPT reflects the models ability to score equally across all question categories, and Harmonic MPT measures the models' ability to have high scores across for harder (low-scoring) question-types. Further, we also report the normalized arithmetic and harmonic MPT, along with traditional overall VQA accuracy.

\textbf{Answer Encoding:} The VQA task is formulated as a classification problem following the benchmark practices \cite{antol2015vqa,Goyal_2017_CVPR,krishna2016visual} where a candidate answer set is created for the most frequent answers in the dataset. This is because VQA datasets have a very long tailed distribution and the least frequent answers account for a fraction of the IQA pairs in the dataset. For experimenting on VQAv2 and VQA-CPv2 datasets, we select the most frequent $3000$ answers and for TDIUC we select $1460$ for the candidate answer set $\mathcal{A}$.

\section{Experiments and Results:}
\label{sec:experiments}
We perform evaluation on the VQAv2, VQA-CPv2 and TDIUC datasets in the scope of this work and group the core experiments into four main categories. 
\begin{itemize}
\item \emph{The Effect of Visual Features:} We vary the input visual feature meta-architecture to evaluate the effect on VQA complexity while using different level of visual feature. 
\item \emph{The Effect of Fusion Meta-architecture:} We vary the fusion meta-architecture to evaluate different so-far proposed strategies and to analyze their complexity-accuracy trade-off while using simpler to complex joint embedding models. 
\item \emph{The Effect of Attention Model:} We further study the effect of additional attention mechanisms on the complexity accuracy trade-off.
\item \emph{Proposed Meta-architecture:} Finally, we find the most effective meta-architecture combination and report state-of-the-art performance on VQAv2, VQA-CPv2 and TDIUC using the recommended meta-architecture combination.  
\end{itemize}

\begin{table*}[!htp]
    \centering
    {
    \begin{tabular}{l cc cc cc cc cc}
    \toprule
    & \multicolumn{10}{c}{Visual Feature} \\\cmidrule(r){2-11}
    Bilinear                & \multicolumn{2}{c}{InceptionV4}   & \multicolumn{2}{c}{ResNet152} & \multicolumn{2}{c}{ResNext101}    & \multicolumn{2}{c}{SeNet154}  & \multicolumn{2}{c}{PolyNet}\\
    Model {$\downarrow$}    & IL & SG                           & IL& SG                        & IL& SG                            &IL & SG                        & IL & SG   \\
    \cmidrule(r){1-1}\cmidrule(r){2-3}\cmidrule(r){4-5}\cmidrule(r){6-7}\cmidrule(r){8-9} \cmidrule(r){10-11}
    Linear  &35.04&36.97    &39.26&39.56    &37.88&38.90    &37.32&38.18    &40.22&38.14\\
    C-MLP   &52.34&54.89    &53.37&58.50    &53.28&57.90    &54.06&57.96    &52.78&56.68\\
    MCB     &52.83&53.44    &54.91&58.15    &55.04&57.94    &55.34&58.23    &55.85&57.29\\
    MLB     &52.66&52.53    &53.79&57.16    &53.77&56.31    &54.69&56.34    &54.91&57.02\\
    Mutan   &53.35&53.97    &55.60&58.94    &55.67&57.21    &55.41&58.11    &55.97&58.75\\
    MFB     &53.88&53.55    &55.47&58.31    &55.45&57.63    &56.16&57.51    &57.69&57.93\\
    Block   &55.08&55.89    &56.85&60.49    &56.87&59.67    &57.36&59.67    &58.12&60.53  \\
    MFH     &54.86&55.28    &57.07&60.53    &57.06&59.89    &57.16&59.64    &57.59&60.53  \\
    \bottomrule
    \end{tabular}
    }
    \caption{Evaluation on VQAv2 \cite{Goyal_2017_CVPR} validation set with visual features extracted using different CNN models.}
    \label{tab:VQAv2_valset}
\end{table*}

We quantify the complexity of a VQA model based on the number of trainable parameters, FLOPS (floating point operations per second) and computation time (both CPU and GPU). The visual features are extracted as a pre-processing step for VQA models; thus pre-training the models on the ImageNet dataset or similar object detection dataset does not directly contribute to the complexity of a VQA model. However, the BU features require training an additional object detector on another large scale dataset (\ie, Visual Genome \cite{krishna2016visual}). Thus we offset the FLOPS and trainable parameters with the additional training cost for performing experiments with BU features, and plot VQA accuracy versus trainable parameter and FLOPS. As our goal is to determine the optimal meta-architecture configuration for highest VQA accuracy, we draw an imaginary \emph{maximum efficiency} line on the accuracy vs. training parameter and accuracy vs. FLOPS plots, that helps us study the overall trends.


\subsection{Varying the level of Visual Features}
\label{sec:vary_vis_feat}

\begin{table*}[!htp]
    \centering
    {
    \begin{tabular}{l cc cc cc cc cc}
    \toprule
    & \multicolumn{10}{c}{Visual Feature} \\\cmidrule(r){2-11}
    Bilinear                & \multicolumn{2}{c}{InceptionV4}   & \multicolumn{2}{c}{ResNet152} & \multicolumn{2}{c}{ResNext101}    & \multicolumn{2}{c}{SeNet154}  & \multicolumn{2}{c}{PolyNet}\\
    Model {$\downarrow$}    & IL & SG                           & IL& SG                        & IL& SG                            &IL & SG                        & IL & SG   \\
    \cmidrule(r){1-1}\cmidrule(r){2-3}\cmidrule(r){4-5}\cmidrule(r){6-7}\cmidrule(r){8-9} \cmidrule(r){10-11}
    Linear  &17.61&17.77    &17.58&19.7  &18.09&19.93  &17.97&29.11  &18.6&25.11  \\
    C-MPL   &27.0&29.23     &27.27&31.38  &27.32&30.23  &28.6&32.31   &26.65&29.35 \\
    MCB     &27.2&28.15     &28.43&30.87  &27.25&29.11  &30.19&31.28  &30.25&31.71 \\
    MLB     &26.1&27.70     &24.61&31.52  &26.13&30.79  &27.87&32.33  &27.6&31.96 \\
    Mutan   &28.25&28.02    &29.27&31.32  &29.64&28.97  &30.75&31.74  &31.04&32.04 \\
    MFB     &27.51&28.69    &28.44&33.05  &28.9&32.38   &30.39&33.61  &29.90&33.32  \\
    Block   &28.45&29.73    &29.17&34.45  &29.41&33.18  &31.0&35.16   &30.71&35.11\\
    MFH     &28.27&30.07    &29.1&34.6  &29.7&34.3  &31.63&35.9   &31.06&35.48  \\
    \bottomrule
    \end{tabular}
    }
    \vspace{-0.5em}
    \caption{Evaluation on VQA-CPv2 \cite{Agrawal_2018_CVPR} test set with visual features extracted using different CNN models.}
    \label{tab:VQA-CP2_testset}
\end{table*}

In Tab.~\ref{tab:VQAv2_valset} we report validation scores on the VQAv2 dataset and in Tab.~\ref{tab:VQA-CP2_testset} we report the test scores on the VQA-CPv2 dataset, using Image Level (IL) and Spatial Grid (SG) features across eight different fusion models. Our main insights are as follows:

\textbf{Setting visual feature dimension closer to language feature embedding improves VQA performance.} It can be seen from Tab.~\ref{tab:VQAv2_valset} and \ref{tab:VQA-CP2_testset} that the VQA models based on InceptionV4 features perform significantly worse (about $\sim 5.0 \downarrow$) than similar models while using IL features instead of SG features (for different CNN backbones). The main reason is that for all our experiments we kept the language feature embedding at $2400$, which is similar to feature dimensions $d_v = 2048$ of other feature extractors. InceptionV4 features have a significantly smaller dimension $d_v = 1536$ than the language feature embedding. While a bilinear fusion model tries to learn a joint feature embedding, the smaller visual feature dimension affects the model's ability to equally capture the visual-semantic relationships, thereby deteriorating VQA accuracy. 

One can use a projection layer to make the visual features high-dimensional (\ie, closer to the dimension of language feature embedding), however, it is generally not recommended. Projecting the visual features to a higher dimension through learned layers introduces a higher complexity and results in over-fitting on a specific dataset. Consequently, the pretrained model does not generalize well to held-out test sets. Another way to make to Inception visual features high dimensional is by increasing the input image size. This approach has practical limitations as we are using a pre-trained feature extraction model with a fixed architecture. In practice, one should not make the visual feature high dimensional to match the language feature embedding, rather one should modify the LSTM architecture to make the semantic feature dimension similar to the visual feature. As the language feature embedding is extracted from the last LSTM cell and is related to the hidden dimension of the cell, it is relatively easy to modify the hidden dimensions to obtain an arbitrary sized language feature embedding.

\textbf{PolyNet features with smaller grid size perform surprisingly well.}
Expect for InceptionV4 and PolyNet, all the other feature extractors we experiment with have $14 \times 14 = 196$ grid locations. Having more grid locations allows a model to learn to identify salient image locations in a higher resolution. However, more grid locations introduce a higher complexity in the VQA models. Surprisingly, PolyNet features with $12 \times 12 = 144$ grid locations, which translates to a $\sim 106$k reduction in visual feature dimension compared to ResNet152 features, achieves the highest VQAv2 validation accuracy while using Block and MFH fusion models (Tab.~\ref{tab:VQAv2_valset}) with SG features. Also, on a more challenging VQA-CPv2 dataset, it performs better then ResNet152 features ($0.7 \uparrow$ and $0.9 \uparrow$ for Block and MFH models, respectively, with SG features). This means that visual features extracted by PolyNet are highly discriminative and can perform on par with higher resolution visual features in a similar VQA setting.

\textbf{Resnet152 features have the largest performance boost when using SG instead of IL features.} In Tab.~\ref{tab:VQAv2_valset} and \ref{tab:VQA-CP2_testset}, we perform experiments using both IL and SG features. While using the SG features, we employ the co-attention mechanism (see component (3) in Fig.~\ref{fig:all_meta_arch}). Naturally, using SG features instead of IL features extracted using the same visual feature extraction meta-architecture results in a significant performance boost. However, the highest improvement is achieved when ResNet152 SG features are used instead of ResNet152 IG features. The average VQA accuracy boost across all fusion models (expect for Linear) is $3.72$ for ResNet152 compared to $0.65, 2.48, 2.47$ and $2.26$ for InceptionV4, ResNext101, SeNet154 and PolyNet, respectively, on the VQAv2 dataset (Tab.~\ref{tab:VQAv2_valset}). The accuracy gain using SG features is more with ResNet152 when we experimented on the VQA-CPv2 dataset; $4.42$ for ResNet152 and $1.25, 2.94, 3.41 $ and $3.11$ for InceptionV4, ResNext101, SeNet154 and PolyNet features, respectively.

\textbf{SeNet154 features perf0rm better on datasets with less language bias.} The VQA-CPv2 dataset allows a more challenging evaluation benchmark for VQA models as its test split has a different schematic data distribution compared to its training counterpart. This prevents a VQA model from cheating by learning the language bias to score higher. In this challenging setting, models using SeNet154 features achieve higher accuracy compared to their performance on the VQAv2 dataset. For example, in Tab.~\ref{tab:VQAv2_valset}, the MFH-SG model with ResNet152 features achieves $60.53$ whereas with SeNet154 it achieves $59.64$. This trend reverses when evaluated on VQA-CPv2 dataset; MFH-SG model with ResNet152 features scores significantly lower ($1.3 \downarrow$) than MFH-SG models using SeNet154 features. This trend is also true for models using SeNet154 IL features.
One possible reason is that SeNet154 has an additional channel attention module that comes in handy when language bias is smaller and the model has to rely on better visual features.

\begin{figure}[t]
\centering
  \includegraphics[width=.8\linewidth]{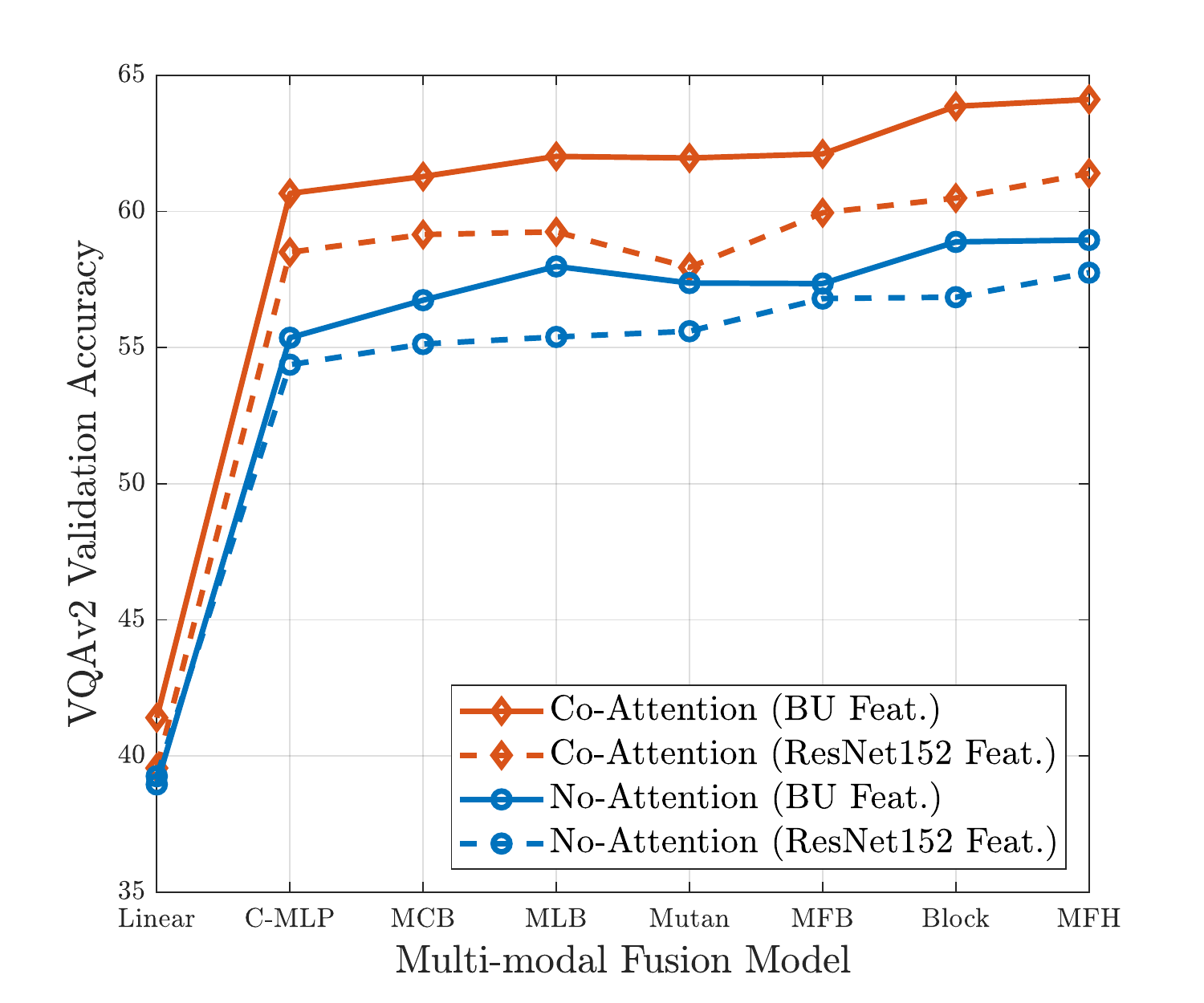}
  \caption{Comparing VQAv2 validation accuracy of Co-Attention and No-Attention version of our Unified VQA model, using ResNet152 Spatial Grid (SG) and Bottom-Up (BU) features.}
\label{fig:vqa2_bu_sg}
\end{figure}

\begin{figure}[t]
\centering
  \includegraphics[width=.8\linewidth]{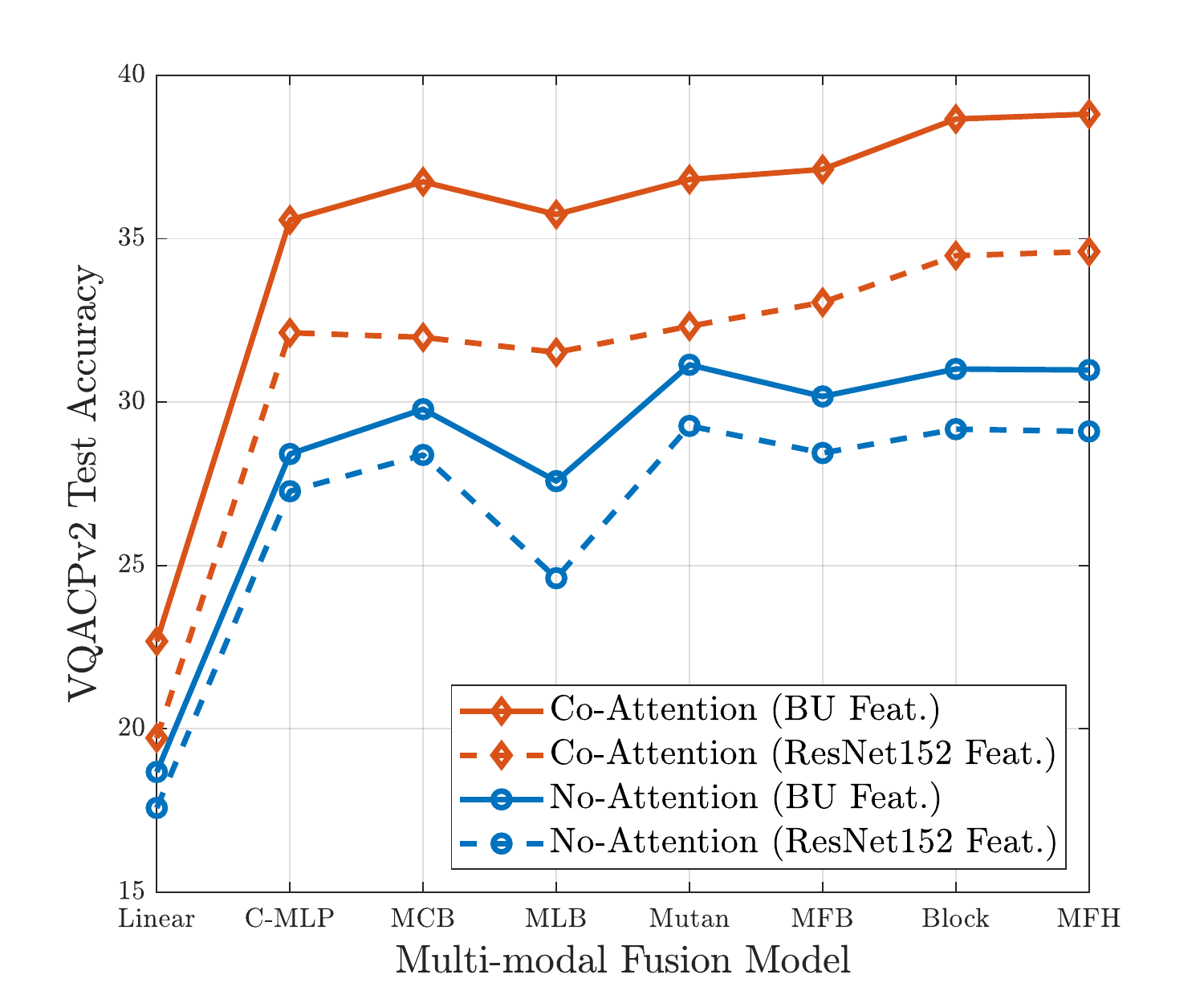}
  \caption{Comparing VQA-CPv2 test accuracy of Co-Attention and No-Attention version of our Unified VQA model, using ResNet152 Spatial Grid (SG) and Bottom-Up (BU) features.}
\label{fig:vqacp2_bu_sg}
\end{figure}

\textbf{Using Bottom-Up features provides  a consistent accuracy gain.} Instead of Spatial Grid features, most state-of-the art VQA models use bottom-up features \cite{Anderson2017up-down}. In this case,  the whole image is represented as a collection of region-based visual features instead of visual features from a fixed grid for every image. As humans naturally tend to ask questions about the objects present in an image, localizing different regions containing objects and their parts allows a VQA model to jump-start the visual attention process and identify which object regions are more important to answer the question. Meanwhile, for a model that takes a uniform grid representation and is required to identify arbitrary image regions relevant to the question. We use the bottom-up features provided by \cite{Anderson2017up-down} for our experiments which uses a ResNet-101 backbone for feature extraction. We compare similar VQA models using BU features with Resnet152 IL and SG features on VQAv2 (Tab.~\ref{fig:vqa2_bu_sg}) and VQA-CPv2(Tab.~\ref{fig:vqacp2_bu_sg}) datasets. We generate Image Level (IL) BU features by average pooling the visual features across the number of objects-proposals to generate $d_v = 2048$ dimensional features. While using no-attention models, we use the image level features and for co-attention model we use spatial-grid/object level features. Form Fig.~\ref{fig:vqa2_bu_sg} and Fig.~\ref{fig:vqacp2_bu_sg} we see that:
\begin{itemize}[leftmargin=*]
  \item[--] \emph{VQA models using BU features perform consistently better than models using ResNet152 features.} On both VQAv2 (Tab.~\ref{fig:vqa2_bu_sg}) and VQA-CPv2 (Tab.~\ref{fig:vqacp2_bu_sg}) datasets, we see that the models using BU features (solid lines) achieve a higher accuracy than the models using ResNet152 features (dashed lines) in all cases. This is because the bottom-features undergo an additional attention step (during top-$N$ object ROI pooling, See Fig.~\ref{fig:vis_feat_meta}) compared to the conventional ResNet features. The no-attention models with BU features have less accuracy gain compared the co-attention models using BU features. This is because the original BU features have $2048$-d visual feature representation for each distinct object, but when they undergo pooling operations, some spatial information pertaining to a single object and its parts is lost. However, when the co-attention models use the BU features, the models learn to generate an attention map over the collection of object proposals which results in a higher accuracy VQA accuracy gain compared to the ResNet152 features.
  
  \item[--] \emph{BU features provide greater accuracy boost in datasets with less language bias.} Comparing Fig.~\ref{fig:vqa2_bu_sg} and Fig.~\ref{fig:vqacp2_bu_sg}, we can see that the accuracy gain from using BU features is greater on the VQA-CPv2 dataset compared to the VQAv2 dataset. The no-attention models on the VQAv2 dataset have an average accuracy gain of $1.3$ when using BU features instead of ResNet features across all fusion models, whereas the gain is $1.8$ on the VQA-CPv2 dataset (Fig.~\ref{fig:vqa2_bu_sg}). This average gain in accuracy is even higher when using the co-attention model, $2.7$ on VQAv2 and $4.1$ on VQA-CPv2 dataset. This is because the BU features encode an additional attention in the form of object proposals whereas ResNet152 or other SG features only provide features uniformly distributed over the spatial grid. The VQA-CPv2 dataset compared to the VQAv2 dataset has less language bias, thus BU features provide a higher accuracy boost in a more challenging setting. 
  \end{itemize}
Even though using BU features instead of ResNet or other SG features improve the VQA accuracy, there is a significant training cost associated with generating BU features which is discussed in more details in the following section (Sec.~\ref{sec:diff_fusion_models}).\\

\begin{figure}[t]
\centering
  \includegraphics[width=.8\linewidth]{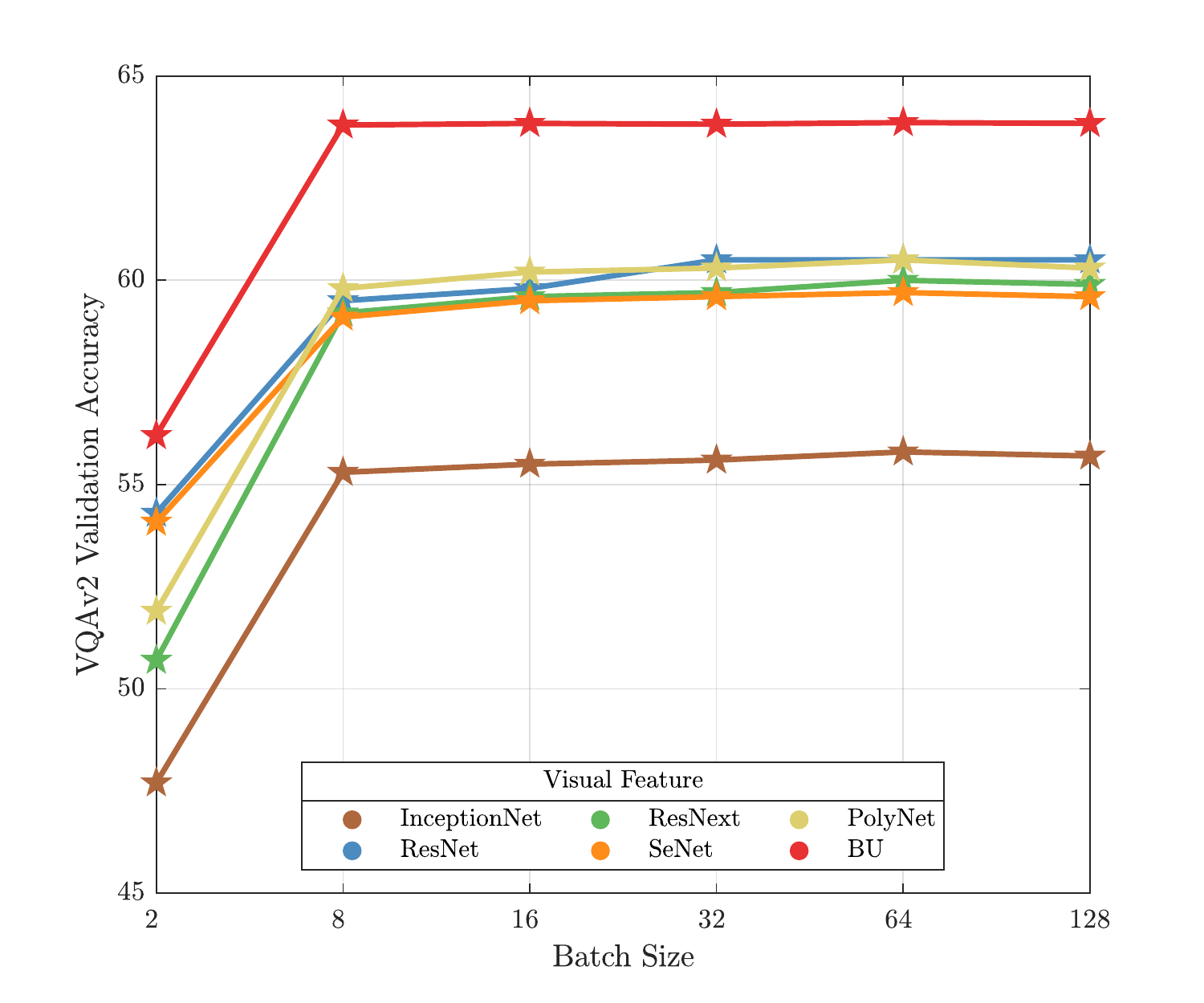}\vspace{-1em}
  \caption{Batch Size vs. VQA accuracy using different CNN backbones used to extract SG features employing Block fusion on VQAv2 validation set.}
\label{fig:vqa_acc_vs_batch_size}
\end{figure}

\textbf{VQA models are less sensitive to change in batch size.} In Fig.~\ref{fig:vqa_acc_vs_batch_size} we report the VQA accuracy of our co-attention model using Block fusion using different CNN extracted SG features and BU features by varying the training batch size from $2$ to $128$. We used a single GPU configuration to perform these experiments for a more robust evaluation and fairer comparison. We see that except for the choice of smallest batch size of 2, the VQA accuracy saturate between $8$ to $128$. For almost all visual feature types, we found the optimal batch size to be $64$. Furthermore, the choice of batch size also depends GPU memory. One can choose a larger batch size and distribute the computational load across multiple GPUs.

\subsection{Employing different fusion models}
\label{sec:diff_fusion_models}
In our analysis, we evaluate the complexity of a VQA model in terms of number of trainable parameters, FLOPS and computation time (both CPU and GPU). The number of trainable parameters in a VQA models mostly depends on the type of fusion models, size of the input visual feature embedding and the candidate answer space. We keep the language feature embedding size and the dimension of candidate answer size fixed for our experiments. The visual feature size varies depending feature extraction meta-architecture which is predefined, and does not contribute to the calculations of trainable parameters or FLOPS except when using BU features. The main variation in the complexity and accuracy calculation comes from the bilinear fusion used in the VQA model. In this section we investigate these VQA accuracy vs. complexity relations by varying the fusion meta-architectures in the VQAv2 and VQA-CPv2 dataset. In this part of the analysis we do not include MCB fusion mechanism as its original implementation was in Caffe \cite{jia2014caffe} which was incompatible our with trainable parameters and FLOPS calculation method.

\begin{table*}[t!]
    \centering
      {
    \begin{tabular}{lcccccccc}
    \toprule
                    &Linear  &C-MLP &MCB   &MLB    &Mutan  &MFB    &Block  &MFH\\
    \midrule
    Scene           &51.0   &92.9   &93.0   &92.6   &92.3  &92.2    &92.8  &92.9\\
    Sport           &19.0   &93.8   &92.8   &93.5   &93.1  &93.5    &93.6  &93.8\\
    Color Att.      &55.7   &68.5   &68.5   &68.6   &66.3  &67.8    &68.7  &67.0\\
    Other Att.      &0.1    &56.4   &56.7   &56.4   &52.1  &57.2    &58.0  &55.9\\
    Activity        &0.0    &52.4   &52.4   &49.0   &49.6  &52.7    &53.2  &51.8\\
    Position        &7.28   &32.2   &35.4   &33.5   &29.4  &32.8    &36.1  &34.7\\
    Sub-Obj         &23.9   &86.1   &85.4   &85.8   &85.8  &85.9    &86.3  &86.1\\
    Absurd          &90.3   &92.5   &84.4   &90.3   &90.0  &93.5    &90.7  &93.3\\
    Utility         &15.2   &26.3   &35.0   &31.6   &27.5  &31.6    &34.5  &35.7\\
    Presence        &93.5   &94.4   &93.6   &93.7   &93.9  &93.7    &94.2  &94.1\\
    Counting        &50.1   &53.1   &51.0   &51.1   &51.3  &51.2    &52.2  &50.7\\
    Sentiment       &56.3   &65.8   &66.3   &64.0   &63.3  &65.3    &66.1  &63.3\\
    \midrule
    AMPT            &38.5   &67.9   &67.9   &67.5   &66.2  &68.1    &68.9  &68.3\\
    HMPT            &0.0    &57.4   &60.5   &58.7   &55.8  &59.2    &61.1  &60.3\\
    N-AMPT          &29.8   &53.8   &42.5   &65.3   &53.6  &53.4    &54.8  &55.6\\
    N-HMPT          &0.0    &28.5   &27.3   &32.2   &32.6  &30.2    &34.2  &38.6\\
    \midrule
    Accuracy        &73.0   &84.0   &81.9   &83.1   &82.7  &83.6    &83.6  &84.3\\
    \bottomrule
    \end{tabular}}\vspace{-0.5em}
    \caption{Evaluation on the testset of TDIUC \cite{kafle2017analysis} dataset with Spatial Grid (SG) ResNet152 features. The first 12 rows report the unnormalized accuracy for each question-type. We report Arithmetic MPT (AMPT) and Harmonic MPT (HMPT) of accuracy scores for all question types alongwith their normalized counterparts N-AMPT and N-HMPT. We also report the traditional VQA accuracy in the last row.}
    \label{tab:TDIUC_testset}
\end{table*}

\textbf{Baseline C-MLP model achieves comparable VQA accuracy in TDIUC and VQAv2 dataset.} For performing evaluation on contemporary VQA datasets, we establish two simple baseline models, namely Linear Summation (Linear) and Concatenation MLP (C-MLP), for a more robust comparison with the state-of-the-art methods. Surprisingly, the simplistic C-MLP model achieves the second highest overall VQA accuracy in the TDIUC testset (Tab.~\ref{tab:TDIUC_testset}) after MFH. It also performs reasonably well in VQAv2 dataset and achieves better VQA accuracy then several state-of-the-art bilinear fusion models. However, in the more challenging VQA-CPv2 dataset, C-MLP performs worse than most of the fusion models. Also, in the TDIUC dataset, the harmonic MPT of C-MLP is less than than other fusion models, because harmonic MPT is skewed towards the question type that has less accuracy. The C-MLP models learns a very high dimensional representation of the image and question distribution through MLP, thus achieves reasonably well in terms of VQA accuracy, but is less generalizable. This means that C-MLP models finds it harder to answer questions requiring superior reasoning capability (\eg, object Utility, relative position), but can easily and more accurately answer question levereging language cues (\eg, color attribute, scene recognition).

\textbf{MFH fusion achieved highest VQA accuracy across all datasets in all settings.} MFH achieves the highest VQA accuracy compared to other fusion models on the VQAv2 validation set, VQA-CPv2 testset and TDIUC dataset. MFH is also consistent when different CNN extracted IL or SG features and BU features are provided. For example, MFH achieves $0.74 \uparrow$ higher accuracy compared to second best fusion model Block while using SeNet154 features on the VQA-CPv2 dataset (Tab.~\ref{tab:VQA-CP2_testset}). Further, it achieved the highest normalized HMPT (N-HMPT) score among all fusion models which means that the MFH bilinear fusion generalizes well across different question types (Tab.~\ref{tab:TDIUC_testset}). 

\subsubsection{Training parameters vs. VQA accuracy}
\begin{figure}[t]
\begin{center}
  \includegraphics[width=0.9\linewidth]{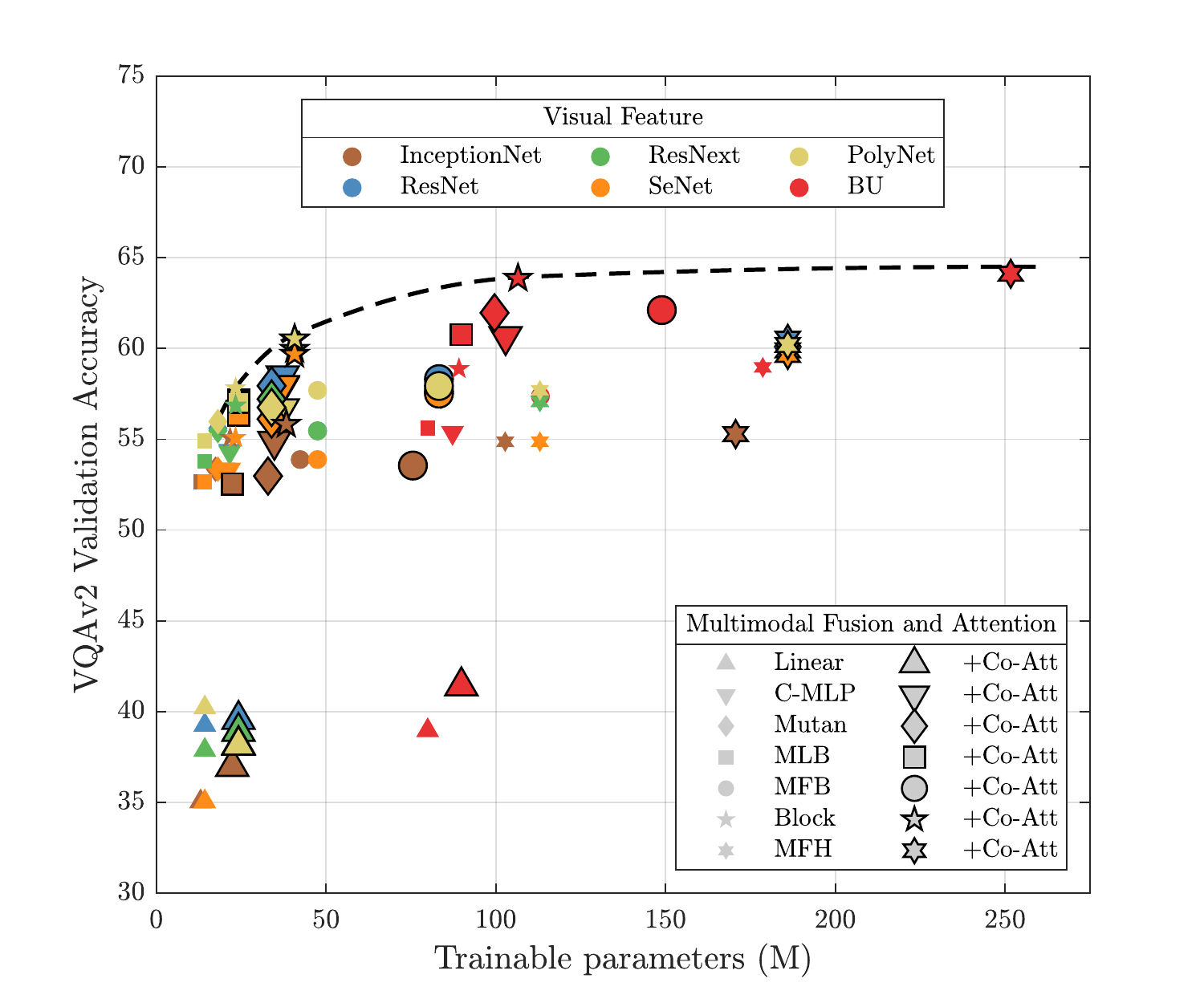}
  \end{center}\vspace{-2em}
  \caption{The trade-off between VQAv2 validation accuracy vs.\ the number of trainable parameters.}
\label{fig:vqa2_acc_vs_param}
\end{figure}
\begin{figure}[t]
\begin{center}
  \includegraphics[width=0.9\linewidth]{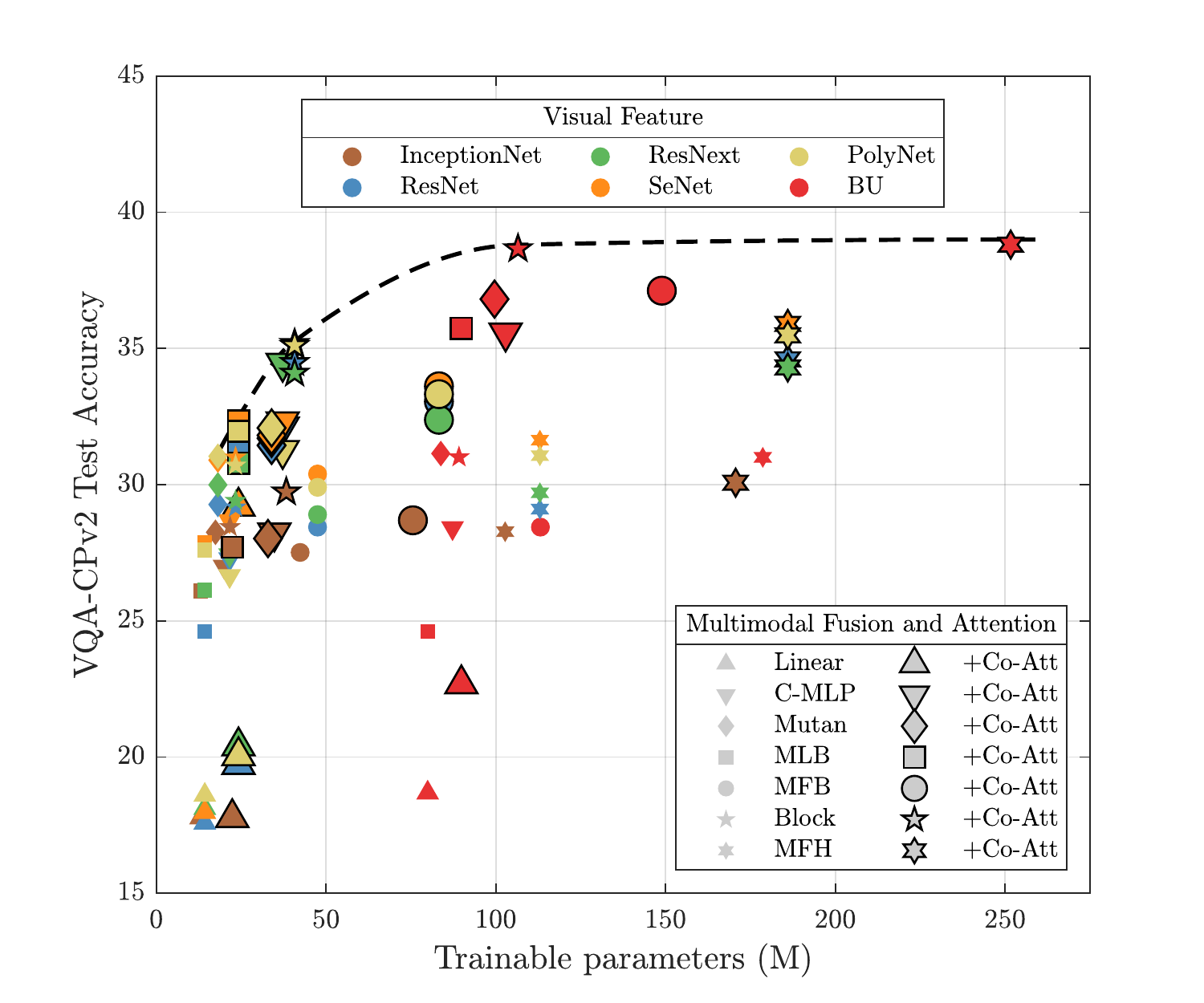}
  \end{center}\vspace{-2em}
  \caption{The trade-off between VQA-CPv2 test accuracy vs.\ the number of trainable parameters.}
\label{fig:vqacp2_acc_vs_param}
\end{figure}
In Fig.~\ref{fig:vqa2_acc_vs_param} and Fig.~\ref{fig:vqacp2_acc_vs_param} we compare VQA accuracy vs. model complexity (number of trainable parameters) respectively on the VQAv2 validation and VQAv2-CP test datasets. Each point in these figures represents a VQA model that was trained on the training set and evaluated on the respective validation/test set. Models employing different visual features are color coded and different fusion strategies are represented by different shapes. The VQA models using a no-attention mechanism are represented with small-sized shapes whereas the models with co-attention mechanism are represented with a larger shape size. Further, in both of the figures, we plot an imaginary \emph{efficiency line} that infers how much better accuracy can be achieved at what additional computational cost.

\textbf{Models using BU features are mostly on the maximum efficiency line at an additional complexity cost.} The VQA models employing superior fusion mechanisms while using BU features achieve the highest accuracy and sit on the maximum efficiency line. Generating the BU features has an additional training cost where a Faster-RCNN \cite{ren2015faster} model pretrained on the ImageNet dataset is again retrained on the Visual Genome dataset \cite{krishna2016visual} with a $1600$ object and $400$ attribute classes. To calculate the additional training cost for generating bottom-up attention, we modified the PyTorch implementation of Faster-RCNN\footnote{\url{https://github.com/jwyang/faster-rcnn.pytorch}} with the additional object and attribute classes, because the original implementation of BU\cite{Anderson2017up-down} was on a legacy version\footnote{\url{https://github.com/peteanderson80/bottom-up-attention}} of Caffe\cite{jia2014caffe}, incompatible with our complexity calculation method. This was done in an effort to 
simulate a realistic training pipeline to estimate the additional training cost. Based on our estimation, we offset the number of trainable parameters of VQA models using BU features by $65.65M$ for a more fairer comparison. Further, as discussed earlier, one can select the $N$ top objects and their $2048$-d visual features as BU features. In our experiments we use $N=36$, but the original BU features have up to $100$ object proposals per image, making the visual feature dimensions even higher which in turn can further increase the number of trainable parameters.

\textbf{VQA models employing the same bilinear fusion are clustered together.} In the VQA accuracy vs. trainable parameters plot we can see the models employing the same fusion mechanism are clustered together and the MFH model has the highest number of trainable parameters. Compared to MFH with BU features, the Block bilinear fusion also performs closer to MFH both on VQAv2 and VQA-CPv2 dataset. Interestingly, we see that the Block model performs significantly worse when using visual features other than BU features compared to similar models using MFH.

\textbf{Models using co-attention achieve a higher VQA accuracy with an added complexity.} As illustrated in Fig.~\ref{fig:all_meta_arch}, the co-attention mechanism has a second bilinear fusion that adds to the overall complexity of the model. However, all the better performing models include the additional a co-attention mechanism, and in VQA-CPv2 (Fig.\ref{fig:vqacp2_acc_vs_param}) the effect of attention is even more prominent than on the VQAv2 dataset (Fig.~\ref{fig:vqa2_acc_vs_param}) with the same number of additional trainable parameters.

\subsubsection{FLOPS vs. VQA accuracy}
\begin{figure}[t]
\begin{center}
  \includegraphics[width=0.9\linewidth]{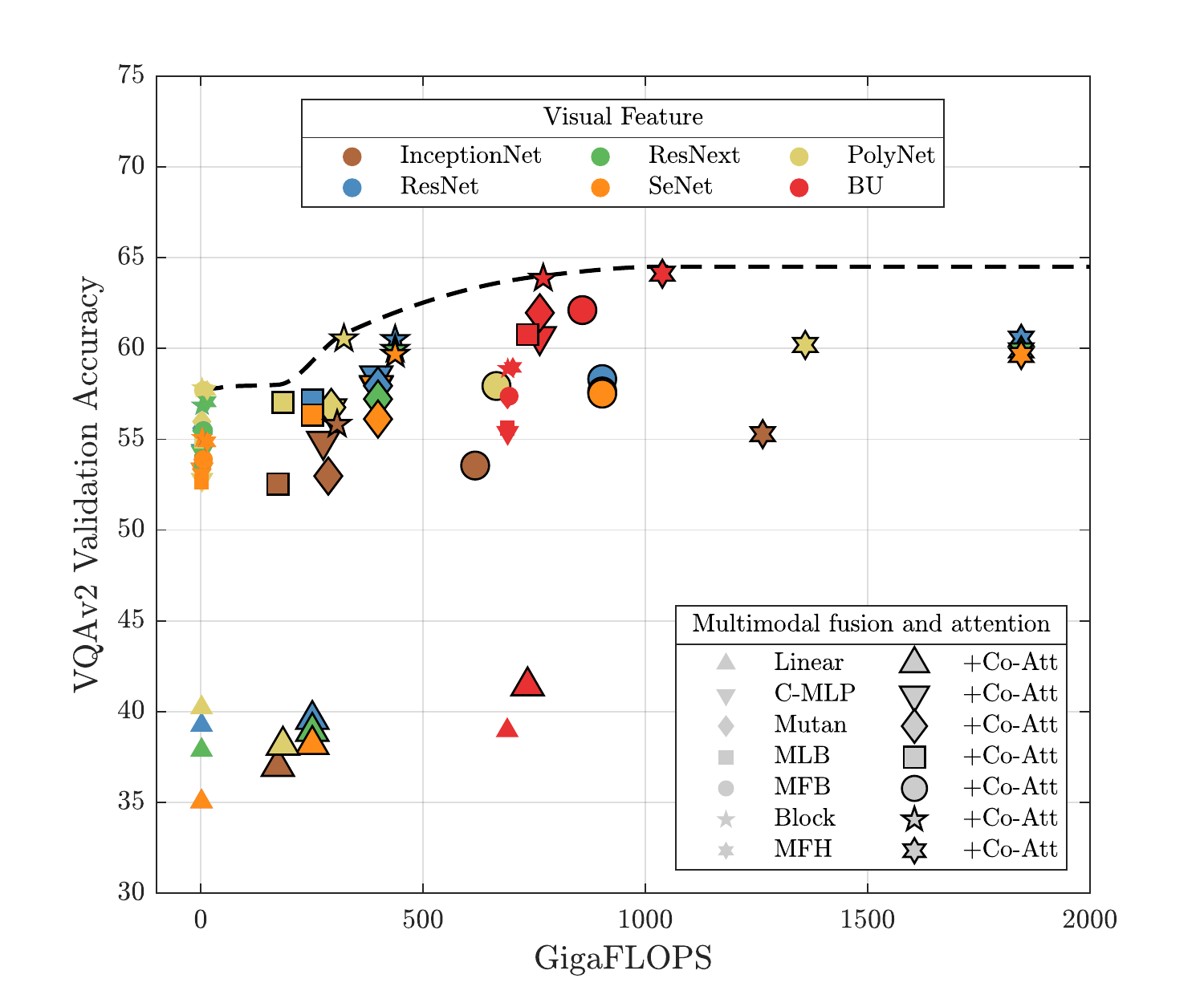}
  \end{center}\vspace{-2em}
  \caption{The trade-off between VQAv2 validation accuracy vs.\ FLOPS.}
\label{fig:vqa2_acc_vs_gflops}
\end{figure}
\begin{figure}[t]
\begin{center}
  \includegraphics[width=0.9\linewidth]{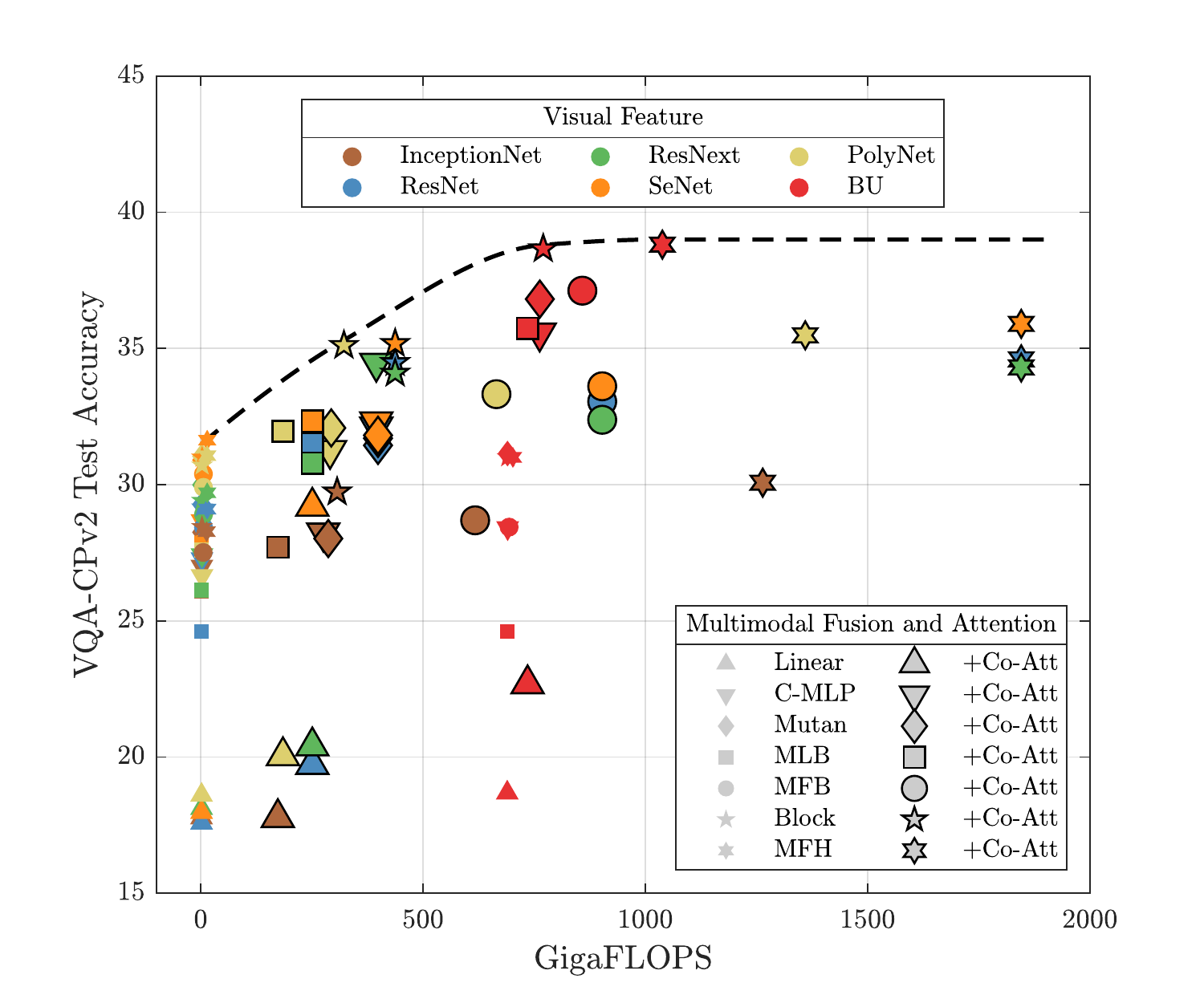}
  \end{center}\vspace{-2em}
  \caption{The trade-off between VQA-CPv2 test accuracy vs. FLOPS.}
\label{fig:vqacp2_acc_vs_gflops}
\end{figure}
In Fig.~\ref{fig:vqa2_acc_vs_gflops} and Fig.~\ref{fig:vqacp2_acc_vs_gflops} we compare VQA accuracy vs. FLOPS (FLoating point OPerations per Second) respectively on the VQAv2 validation and VQAv2-CP test dataset. We use Thop\footnote{\url{https://github.com/Lyken17/pytorch-OpCounter}} to calculate the number of FLOPS. Similar to offsetting the number of trainable parameters while using BU features, we include an offset of $687$ Giga-FLOPS to generate $36$ object proposals and the associated BU features. Below, we summarize the key findings. 

\textbf{Processing SG features requires more FLOPS compared to BU features.} For our experiments, the BU features are $36 \times 2048$ dimensional and the SG features, such as from ResNet152, are $196 \times 2048$ dimensional. The visual feature dimension is directly proportional to the FLOPS. Specifically, as the visual feature dimension increases, the FLOPS count increases exponentially.

\textbf{MFH model requires the highest number of FLOPS.} As expected given the higher number of training parameters, the MFH bilinear fusion based VQA models require the largest number of FLOPS compared to other joint embedding models. The second highest FLOPS count is that of MFB fusion approach followed by the Block. Simplistic fusion strategies such as Linear and C-MLP require the least number of FLOPS.

\subsubsection{Computation time vs. VQA accuracy}
\begin{figure}[t]
\begin{center}
  \includegraphics[width=.8\linewidth]{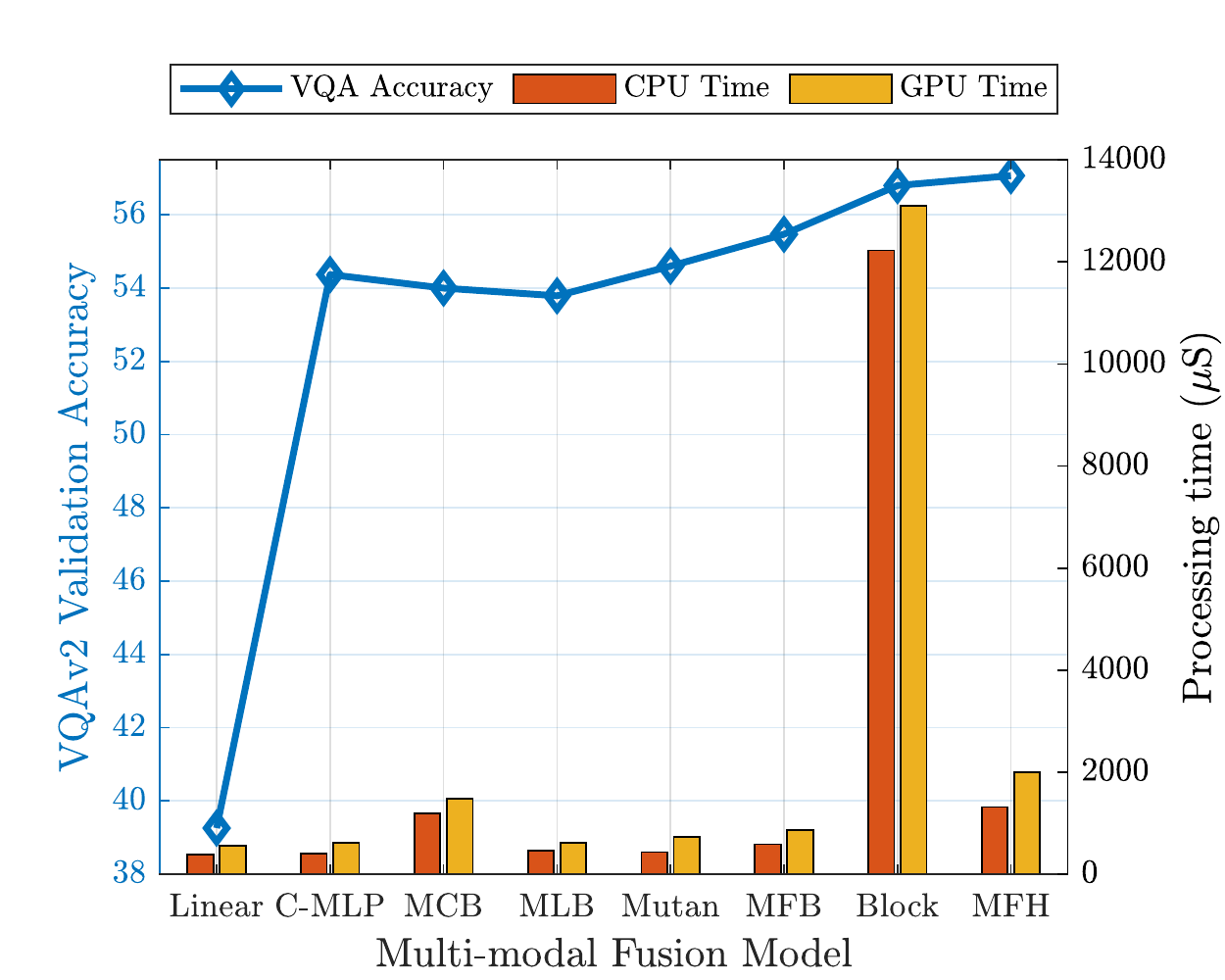}
  \end{center}\vspace{-1.5em}
  \caption{Computation time (CPU and GPU) while employing ResNet152 image level (IL) features with different fusion models.}
\label{fig:cpu_gpu_time}
\end{figure}
We use \texttt{torch.autograd.profile} to report the CPU and GPU times. We use a no-attention model in this experiment because we only want to know which joint embedding model is faster/slower in a comparable setting and the trend we find stays applicable for the co-attention based models. We set the batch size at 64 with one Tesla P100 SXM2 16GB GPU and report the average computation time of 10 mini-batches, each containing 10 Image-Question-Answer (IQA) triplets during  the training time. We perform the evaluation on the VQAv2 validation dataset using ResNet152 features ($d_v = 2048 \times 14 \times 14$) and report our findings in Fig.~\ref{fig:cpu_gpu_time}. The left x-axis of Fig.~\ref{fig:cpu_gpu_time} represent the VQA accuracy and the right x-axis represents the computation time in micro-seconds ($\mu S$). We do not factor the time for I/O operations as it might vary arbitrarily depending on the system configuration. We use the same system configuration for all our experiments so that the result is not biased by the I/O operation. 

\textbf{Block fusion model takes a significantly longer time than other fusion models.} The Block fusion model achieves second-best VQA accuracy but requires a significantly longer time than other fusion models. This is because the Block model decomposes the core tensor into multiple blocks/chunks which separately embed a fraction of the input feature representation in the joint embedding space, and the computational time exponentially increases when the number of blocks increase. On the other hand, even though MFH has more trainable parameters, it achieves a higher VQA accuracy with a fraction of CPU and GPU time compared to Block model.

\subsection{Effect of Co-attention meta-architecture}

\textbf{Adding Co-attention results in more VQA accuracy gain on the challenging VQAv2-CP dataset.} Throughout our experiments we found that co-attention mechanism improves VQA accuracy at the cost of some additional complexity. With the co-attention mechanism, a VQA model is able to learn a question-specific attention distribution over the image and its parts, which is more important when experimenting on the VQA-CPv2 dataset. From Tab.~\ref{tab:VQAv2_valset} and Tab.~\ref{tab:VQA-CP2_testset}, we see that across all fusion models (except for Linear) and visual features, the average accuracy gain with co-attention mechanism on the VQAv2 validation set is $2.32$ and on the VQA-CPv2 test set the gain is $3.02$. This suggests that a superior attention mechanism is a necessary requisite when the experimentation dataset is more challenging and requires intelligent reasoning.

\textbf{Linear model with co-attention achieves $\mathbf{11\times}$ accuracy gain on the VQA-CPv2 dataset, compared to the VQAv2 dataset with SeNet154 features.} With the Linear baseline model, we simply sum up the linear projections of the input feature and project them back to the answer embedding space. Interestingly, when the Linear model is used with co-attention, the accuracy gain is very low ($0.4$ across all CNN extracted features) on the VQAv2 validation dataset compared to other fusion models, even sometimes negative. For example, with the Linear model using PolyNet features, the accuracy drops by $2.08$ when co-attention is used (Tab.~\ref{tab:VQAv2_valset}). On the other hand, when experimenting on the VQA-CPv2 dataset, Linear models with co-attention report high VQA accuracy gain, $11.14$ and $6.51$ while using SeNet154 features and PolyNet features respectively, with an average accuracy gain of $4.5$ (more than $11\times$ increase) across all CNN extracted visual features (Tab.~\ref{tab:VQA-CP2_testset}). 

\subsection{Proposed meta-architecture recommendation}
\label{sec:recommendation}
We did an extensive evaluation of different meta-architectures to study the accuracy vs. complexity trade-off for VQA models. We recommend two settings based on our evaluation. \emph{First}, a less computationally expensive setting that achieves reasonable performance with faster training and inference time. Further, we recommend a \emph{second} setting, that achieves state-of-the-art performance on VQAv2, VQA-CPv2 and TDIUC datasets. 

\subsubsection{Low complexity setting} 
From our analysis, we see that using BU features yield a better VQA accuracy but have significant additional training cost. Further, incorporating BU features in an end-to-end setting can be challenging and is less generalizable. Thus, for a low complexity setting we recommend using CNN extracted SG features with co-attention, specifically the SeNet features. SeNet encodes additional channel attention, thereby the visual features are more discriminative and have the same feature dimension as popular ResNet features. For the fusion model, we recommend to use the C-MLP model because it performs very close to the state-of-the-art on three benchmark VQA datasets and is lightweight. Further, C-MLP is comparatively easy to implement and can be modified to increase or decrease the complexity of the model by simply changing the hidden dimensions of the MLP layers. This would allow a practical VQA setup flexibility with reasonable VQA performance.

\subsubsection{High VQA accuracy setting}
For a VQA model to achieve the best accuracy, we recommend using pre-processed BU features with attention and the MFH bilinear model to jointly embed visual and semantic features. Our dataset-wise results are given below.

\textbf{TDIUC dataset:} Without modifying our Unified VQA model (as in Fig.~\ref{fig:all_meta_arch}), using MFH with BU features and co-attention, we achieved state-of-the-art performance on the TDIUC dataset. We report the accuracy of our U-VQA model against other state-of-the-art methods in Tab.~\ref{tab:TDIUC_SOTA}
\begin{table*}[ht!]
    \centering
    {
    \begin{tabular}{lccccc}
    \toprule
    Model   & Accuracy & AMPT & HMPT & N-AMPT & N-HMPT\\
    \midrule
    NMN \cite{andreas2016neural}
    &79.56  &62.59  &51.87  &34.00  &16.67\\
    MCB \cite{fukui2016multimodal}
    &81.86  &67.90  &60.47  &42.24  &27.28\\
    RAU \cite{noh2016training}
    &84.26  &67.81  &59.99  &41.04  &23.99\\
    QAS \cite{shi2018question}
    &85.03  &69.11  &60.08  &-  &-\\
    Block \cite{ben2019block}
    &85.96  & 71.84 &65.52 &58.36 &39.44 \\
    \midrule
    Ours(U-VQA+MFH) 
    &\textbf{86.33} &\textbf{81.54} &\textbf{65.81}   &\textbf{58.80}   &\textbf{41.76}\\
    \bottomrule
    \end{tabular}}
    \caption{Comparison with state-of-the-art methods on TDIUC \cite{kafle2017analysis} testset.}
    \label{tab:TDIUC_SOTA}
\end{table*}

\textbf{VQAv2 dataset:} MCAN \cite{yu2019deep} currently achieves the state-of-the-art performance on VQAv2 dataset by employing a deep modular co-attention mechanism.  Throughout our experiments, we found out that an additional attention mechanism can help VQA models achieve better accuracy. For achieving higher accuracy than MCAN on VQAv2 dataset, we adopt their implementation of deep modular co-attention\footnote{https://github.com/MILVLG/openvqa/tree/master/openvqa/models/mcan}. They use a linear multimodal fusion operation to jointly embed attended image and question features before the classification layer. As we found MFH to be a superior bilinear fusion model, we replace the linear fusion operation with MFH and use BU features for our experiments. Even though, MCAN is a highly engineered setup that achieves state-of-the-art performance on the saturated VQAv2 dataset, following our meta-architecture recommendations, we report overall VQA accuracy improvements of $0.13$ and  $0.18$ respectively on VQAv2 test-dev and test-std datasets (Tab.~\ref{tab:VQAv2_SOTA}).

\begin{table}[ht!]
    \centering
    {
    \begin{tabular}{lccccc}
    \toprule
            &\multicolumn{4}{c}{Test-dev}           & Test-std\\
            \cmidrule(lr){2-5} \cmidrule(lr){6-6}
    Model   &Overall    &Y/N    &Number    &Other   & Overall\\
    \midrule
    MCB\cite{fukui2016multimodal}
    &62.27  &78.46  &38.28  &57.80  &53.36\\
    BU \cite{Anderson2017up-down}
    &65.32  &81.82  &44.21  &56.05  &65.67\\
    Mutan\cite{benyounescadene2017mutan}$^\dagger$
    &66.01  &82.88  &44.54  &56.50  &66.38\\
    MLB\cite{kim2016hadamard}$^\dagger$
    &66.27  &83.58  &44.92  &56.34  &66.62\\
    RAF\cite{Farazi_2018_BMVC}
    &67.20  &84.10  &44.90  &57.80  &67.40\\ 
    Block\cite{ben2019block}
    &67.58  &83.60  &47.33  &58.51  &67.92\\
    MuRel\cite{cadene2019murel}
    &68.03  &84.77  &49.84  &57.85  &68.41\\
    Counter\cite{zhang2018vqacount}
    &68.09  &83.14  &51.62  &58.97  &68.41\\
    MFH \cite{yu2018beyond}
    &68.76  &84.27  &49.56  &59.89  &-\\
    BAN\cite{kim2018bilinear}
    &69.52  &85.31  &50.93  &60.26  &-\\
    BAN+Counter\cite{kim2018bilinear}
    &70.04  &85.42  &\textbf{54.04}  &60.52  &70.35\\
    MCAN\cite{yu2019deep}
    &70.63  &86.82  &53.26  &60.72  &70.90\\
    \midrule 
    Ours (MCAN+MFH)
    &\textbf{70.76}  &\textbf{87.1}   &53.21  &\textbf{60.77} &\textbf{71.08}\\
    \bottomrule
    \end{tabular}}
    \caption{Comparison with state-of-the-art methods on VQAv2 \cite{Agrawal_2018_CVPR} test-dev and test-std dataset. ($\dagger$) reproted from \cite{cadene2019murel}.}
    \label{tab:VQAv2_SOTA}
\end{table}

\textbf{VQA-CPv2 dataset:} Similar to our approach on the VQAv2 dataset, we found that RUBi \cite{cadene2019rubi} currently achieves that state-of-the-art performance by adding an additional question only branch that reduces the language bias inherent to the dataset. This approach is particularly useful on the VQA-CPv2 dataset since by its design the train and test splits have different semantic distribution. In their baseline architecture, they use a Block fusion model to jointly embed visual and semantic features. We replace the Block fusion model in their baseline architecture\footnote{https://github.com/cdancette/rubi.bootstrap.pytorch} with MFH and report a $1.33$ accuracy gain over the current state-of-the-art (Tab.~\ref{tab:VQA-CP2_SOTA}).

\begin{table}[ht!]
    \centering
     {
    \begin{tabular}{lcccc}
    \toprule
    Model   &Overall    &Y/N    &Number    &Other\\
    \midrule
    VQA \cite{antol2015vqa}
    &19.73  &34.25  &11.39  &14.41\\
    NMN \cite{andreas2016neural}
    &27.47  &38.94  &11.92  &25.72\\
    MCB \cite{fukui2016multimodal}
    &36.33  &41.01  &11.96  &40.57\\
    GVQA \cite{Agrawal_2018_CVPR}
    &31.30  &57.99  &13.68  &22.14\\
    MuRel \cite{cadene2019murel}
    &39.54  &42.85  &13.17  &45.04\\
    Q-Adv \cite{ramakrishnan2018overcoming}
    &41.71  &64.49  &15.48  &35.48\\
    RUBi \cite{cadene2019rubi}
    &47.11  &68.65  &20.28  &43.18\\
    \midrule
    Ours (RUBi+MFH)
    &\textbf{48.44}  &\textbf{73.04}  &\textbf{21.43}  &\textbf{43.44}\\
    \bottomrule
    \end{tabular}}
    \vspace{-0.5em}
    \caption{\small{Comparison with state-of-the-art methods on VQA-CPv2 \cite{Agrawal_2018_CVPR} testset.}}
    \label{tab:VQA-CP2_SOTA}
\end{table}

\section{Conclusion}
\label{sec:conclusion}
Visual question answering (VQA) is a challenging problem that is actively under investigation. A range of existing approaches exist in the literature, all developed with different ingredients, that makes it difficult to make a fair comparison between them. In this work, we systematically study the influence of key components commonly used within VQA models on the efficiency and final performance. We performed extensive evaluation on three benchmark VQA datasets by varying the VQA meta-architecture. Based on our extensive experiments, we provide two recommendations for meta-architecture selection. One focuses on achieving reasonable VQA accuracy with a simple and light weight architecture, while the other focuses on achieving the state-of-the-art accuracy on VQAv2, VQA-CPv2 and TDIUC datasets. We hope that our findings and recommendations will help researchers to find optimum design choices for VQA and other multi-modal tasks based on vision and language inputs. 

\bibliography{egbib}

\begin{thebibliography}{10}
\expandafter\ifx\csname url\endcsname\relax
  \def\url#1{\texttt{#1}}\fi
\expandafter\ifx\csname urlprefix\endcsname\relax\def\urlprefix{URL }\fi
\expandafter\ifx\csname href\endcsname\relax
  \def\href#1#2{#2} \def\path#1{#1}\fi

\bibitem{antol2015vqa}
S.~Antol, A.~Agrawal, J.~Lu, M.~Mitchell, D.~Batra, C.~Lawrence~Zitnick,
  D.~Parikh, Vqa: Visual question answering, in: Proceedings of the IEEE
  International Conference on Computer Vision, 2015, pp. 2425--2433.

\bibitem{gu2018recent}
J.~Gu, Z.~Wang, J.~Kuen, L.~Ma, A.~Shahroudy, B.~Shuai, T.~Liu, X.~Wang,
  G.~Wang, J.~Cai, et~al., Recent advances in convolutional neural networks,
  Pattern Recognition 77 (2018) 354--377.

\bibitem{simonyan2014very}
K.~Simonyan, A.~Zisserman, Very deep convolutional networks for large-scale
  image recognition, arXiv preprint arXiv:1409.1556.

\bibitem{he2016deep}
K.~He, X.~Zhang, S.~Ren, J.~Sun, Deep residual learning for image recognition,
  in: Proceedings of the IEEE Conference on Computer Vision and Pattern
  Recognition, 2016, pp. 770--778.

\bibitem{xie2017aggregated}
S.~Xie, R.~Girshick, P.~Doll{\'a}r, Z.~Tu, K.~He, Aggregated residual
  transformations for deep neural networks, in: Proceedings of the IEEE
  conference on computer vision and pattern recognition, 2017, pp. 1492--1500.

\bibitem{fukui2016multimodal}
A.~Fukui, D.~H. Park, D.~Yang, A.~Rohrbach, T.~Darrell, M.~Rohrbach, Multimodal
  compact bilinear pooling for visual question answering and visual grounding,
  arXiv preprint arXiv:1606.01847.

\bibitem{kiros2015skip}
R.~Kiros, Y.~Zhu, R.~R. Salakhutdinov, R.~Zemel, R.~Urtasun, A.~Torralba,
  S.~Fidler, Skip-thought vectors, in: Advances in neural information
  processing systems, 2015, pp. 3294--3302.

\bibitem{krishna2016visual}
R.~Krishna, Y.~Zhu, O.~Groth, J.~Johnson, K.~Hata, J.~Kravitz, S.~Chen,
  Y.~Kalantidis, L.-J. Li, D.~A. Shamma, et~al., Visual genome: Connecting
  language and vision using crowdsourced dense image annotations, arXiv
  preprint arXiv:1602.07332.

\bibitem{zhu2016visual7w}
Y.~Zhu, O.~Groth, M.~Bernstein, L.~Fei-Fei, Visual7w: Grounded question
  answering in images, in: Proceedings of the IEEE Conference on Computer
  Vision and Pattern Recognition, 2016, pp. 4995--5004.

\bibitem{benyounescadene2017mutan}
H.~Ben-Younes, R.~Cad{\`{e}}ne, N.~Thome, M.~Cord, Mutan: Multimodal tucker
  fusion for visual question answering, ICCV.

\bibitem{ben2019block}
H.~Ben-Younes, R.~Cadene, N.~Thome, M.~Cord, Block: Bilinear superdiagonal
  fusion for visual question answering and visual relationship detection, in:
  AAAI 2019-33rd AAAI Conference on Artificial Intelligence, 2019.

\bibitem{lu2016hierarchical}
J.~Lu, J.~Yang, D.~Batra, D.~Parikh, Hierarchical question-image co-attention
  for visual question answering, in: Advances In Neural Information Processing
  Systems, 2016, pp. 289--297.

\bibitem{yang2016stacked}
Z.~Yang, X.~He, J.~Gao, L.~Deng, A.~Smola, Stacked attention networks for image
  question answering, in: Proceedings of the IEEE Conference on Computer Vision
  and Pattern Recognition, 2016, pp. 21--29.

\bibitem{jabri2016revisiting}
A.~Jabri, A.~Joulin, L.~van~der Maaten, Revisiting visual question answering
  baselines, in: European Conference on Computer Vision, Springer, 2016, pp.
  727--739.

\bibitem{xu2015show}
K.~Xu, J.~Ba, R.~Kiros, K.~Cho, A.~Courville, R.~Salakhudinov, R.~Zemel,
  Y.~Bengio, Show, attend and tell: Neural image caption generation with visual
  attention, in: International Conference on Machine Learning, 2015, pp.
  2048--2057.

\bibitem{kim2018bilinear}
J.-H. Kim, J.~Jun, B.-T. Zhang, Bilinear attention networks, in: Advances in
  Neural Information Processing Systems, 2018, pp. 1564--1574.

\bibitem{hu2018squeeze}
J.~Hu, L.~Shen, G.~Sun, Squeeze-and-excitation networks, in: Proceedings of the
  IEEE conference on computer vision and pattern recognition, 2018, pp.
  7132--7141.

\bibitem{Anderson2017up-down}
P.~Anderson, X.~He, C.~Buehler, D.~Teney, M.~Johnson, S.~Gould, L.~Zhang,
  Bottom-up and top-down attention for image captioning and visual question
  answering, in: CVPR, 2018.

\bibitem{ren2015faster}
S.~Ren, K.~He, R.~Girshick, J.~Sun, Faster r-cnn: Towards real-time object
  detection with region proposal networks, in: Advances in neural information
  processing systems, 2015, pp. 91--99.

\bibitem{pennington2014glove}
J.~Pennington, R.~Socher, C.~Manning, Glove: Global vectors for word
  representation, in: Proceedings of the 2014 conference on empirical methods
  in natural language processing (EMNLP), 2014, pp. 1532--1543.

\bibitem{hochreiter1997long}
S.~Hochreiter, J.~Schmidhuber, Long short-term memory, Neural computation 9~(8)
  (1997) 1735--1780.

\bibitem{devlin2018bert}
J.~Devlin, M.-W. Chang, K.~Lee, K.~Toutanova, Bert: Pre-training of deep
  bidirectional transformers for language understanding, arXiv preprint
  arXiv:1810.04805.

\bibitem{charikar2002finding}
M.~Charikar, K.~Chen, M.~Farach-Colton, Finding frequent items in data streams,
  in: International Colloquium on Automata, Languages, and Programming,
  Springer, 2002, pp. 693--703.

\bibitem{kim2016hadamard}
J.-H. Kim, K.-W. On, J.~Kim, J.-W. Ha, B.-T. Zhang, Hadamard product for
  low-rank bilinear pooling, arXiv preprint arXiv:1610.04325.

\bibitem{yu2018beyond}
Z.~Yu, J.~Yu, C.~Xiang, J.~Fan, D.~Tao, Beyond bilinear: Generalized multimodal
  factorized high-order pooling for visual question answering, IEEE
  Transactions on Neural Networks and Learning Systems.

\bibitem{tucker1966some}
L.~R. Tucker, Some mathematical notes on three-mode factor analysis,
  Psychometrika 31~(3) (1966) 279--311.

\bibitem{yu2017multi}
D.~Yu, J.~Fu, T.~Mei, Y.~Rui, Multi-level attention networks for visual
  question answering, in: Conf. on Computer Vision and Pattern Recognition,
  2017.

\bibitem{szegedy2015going}
C.~Szegedy, W.~Liu, Y.~Jia, P.~Sermanet, S.~Reed, D.~Anguelov, D.~Erhan,
  V.~Vanhoucke, A.~Rabinovich, Going deeper with convolutions, in: Proceedings
  of the IEEE conference on computer vision and pattern recognition, 2015, pp.
  1--9.

\bibitem{zhang2017polynet}
X.~Zhang, Z.~Li, C.~Change~Loy, D.~Lin, Polynet: A pursuit of structural
  diversity in very deep networks, in: Proceedings of the IEEE Conference on
  Computer Vision and Pattern Recognition, 2017, pp. 718--726.

\bibitem{NEURIPS2019_9015}
A.~Paszke, S.~Gross, F.~Massa, A.~Lerer, J.~Bradbury, G.~Chanan, T.~Killeen,
  Z.~Lin, N.~Gimelshein, L.~Antiga, A.~Desmaison, A.~Kopf, E.~Yang, Z.~DeVito,
  M.~Raison, A.~Tejani, S.~Chilamkurthy, B.~Steiner, L.~Fang, J.~Bai,
  S.~Chintala, Pytorch: An imperative style, high-performance deep learning
  library, in: H.~Wallach, H.~Larochelle, A.~Beygelzimer, F.~d\textquotesingle
  Alch\'{e}-Buc, E.~Fox, R.~Garnett (Eds.), Advances in Neural Information
  Processing Systems 32, Curran Associates, Inc., 2019, pp. 8024--8035.

\bibitem{Goyal_2017_CVPR}
Y.~Goyal, T.~Khot, D.~Summers-Stay, D.~Batra, D.~Parikh, Making the v in vqa
  matter: Elevating the role of image understanding in visual question
  answering, in: The IEEE Conference on Computer Vision and Pattern Recognition
  (CVPR), 2017.

\bibitem{Agrawal_2018_CVPR}
A.~Agrawal, D.~Batra, D.~Parikh, A.~Kembhavi, Don't just assume; look and
  answer: Overcoming priors for visual question answering, in: The IEEE
  Conference on Computer Vision and Pattern Recognition (CVPR), 2018.

\bibitem{kafle2017analysis}
K.~Kafle, C.~Kanan, An analysis of visual question answering algorithms, in:
  Proceedings of the IEEE International Conference on Computer Vision, 2017,
  pp. 1965--1973.

\bibitem{lin2014microsoft}
T.-Y. Lin, M.~Maire, S.~Belongie, J.~Hays, P.~Perona, D.~Ramanan,
  P.~Doll{\'a}r, C.~L. Zitnick, Microsoft coco: Common objects in context, in:
  European Conference on Computer Vision, Springer, 2014, pp. 740--755.

\bibitem{jia2014caffe}
Y.~Jia, E.~Shelhamer, J.~Donahue, S.~Karayev, J.~Long, R.~Girshick,
  S.~Guadarrama, T.~Darrell, Caffe: Convolutional architecture for fast feature
  embedding, arXiv preprint arXiv:1408.5093.

\bibitem{andreas2016neural}
J.~Andreas, M.~Rohrbach, T.~Darrell, D.~Klein, Neural module networks, in:
  Proceedings of the IEEE Conference on Computer Vision and Pattern
  Recognition, 2016, pp. 39--48.

\bibitem{noh2016training}
H.~Noh, B.~Han, Training recurrent answering units with joint loss minimization
  for vqa, arXiv preprint arXiv:1606.03647.

\bibitem{shi2018question}
Y.~Shi, T.~Furlanello, S.~Zha, A.~Anandkumar, Question type guided attention in
  visual question answering, in: Proceedings of the European Conference on
  Computer Vision (ECCV), 2018, pp. 151--166.

\bibitem{yu2019deep}
Z.~Yu, J.~Yu, Y.~Cui, D.~Tao, Q.~Tian, Deep modular co-attention networks for
  visual question answering, in: Proceedings of the IEEE Conference on Computer
  Vision and Pattern Recognition, 2019, pp. 6281--6290.

\bibitem{Farazi_2018_BMVC}
M.~R. Farazi, S.~Khan, Reciprocal attention fusion for visual question
  answering, in: The British Machine Vision Conference (BMVC), 2018.

\bibitem{cadene2019murel}
R.~Cadene, H.~Ben-Younes, M.~Cord, N.~Thome, Murel: Multimodal relational
  reasoning for visual question answering, in: Proceedings of the IEEE
  Conference on Computer Vision and Pattern Recognition, 2019, pp. 1989--1998.

\bibitem{zhang2018vqacount}
Y.~Zhang, J.~Hare, A.~Pr\"ugel-Bennett, Learning to count objects in natural
  images for visual question answering, in: International Conference on
  Learning Representations, 2018.
\newblock \href {http://arxiv.org/abs/1802.05766} {\path{arXiv:1802.05766}}.

\bibitem{cadene2019rubi}
R.~Cadene, C.~Dancette, H.~Ben-younes, M.~Cord, D.~Parikh, Rubi: Reducing
  unimodal biases in visual question answering, arXiv preprint
  arXiv:1906.10169.

\bibitem{ramakrishnan2018overcoming}
S.~Ramakrishnan, A.~Agrawal, S.~Lee, Overcoming language priors in visual
  question answering with adversarial regularization, in: Advances in Neural
  Information Processing Systems, 2018, pp. 1541--1551.

\end{thebibliography}

\end{document}